\PassOptionsToPackage{table}{xcolor}
\documentclass{article} 
\usepackage{iclr2027_conference,times}

\usepackage{amsmath,amsfonts,bm}

\def\eqref#1{equation~\ref{#1}}

\def\1{\bm{1}}

\DeclareMathAlphabet{\mathsfit}{\encodingdefault}{\sfdefault}{m}{sl}
\SetMathAlphabet{\mathsfit}{bold}{\encodingdefault}{\sfdefault}{bx}{n}

\usepackage{graphicx}
\usepackage{xcolor}
\usepackage{hyperref}
\usepackage{url}
\usepackage{amsmath}
\usepackage{cleveref}
\crefname{figure}{Fig.}{Figs.}
\Crefname{figure}{Fig.}{Figs.}
\crefname{section}{Sec.}{Secs.}
\Crefname{section}{Sec.}{Secs.}
\crefname{subsection}{Sec.}{Secs.}
\Crefname{subsection}{Sec.}{Secs.}
\crefname{subsubsection}{Sec.}{Secs.}
\Crefname{subsubsection}{Sec.}{Secs.}
\crefname{appendix}{Appendix}{Appendices}
\Crefname{appendix}{Appendix}{Appendices}
\crefname{table}{Table}{Tables}
\Crefname{table}{Table}{Tables}
\usepackage{booktabs}
\usepackage{subcaption}  
\usepackage{makecell}
\usepackage{multirow}
\usepackage{amssymb}
\usepackage{tabularx}
\usepackage{wrapfig}
\usepackage{placeins}
\usepackage{fvextra}
\newcommand{\NR}{\textsc{NR}}

\usepackage{arydshln}

\title{Why Full-Duplex Speech Models Collapse in Long Conversations}

\author{
Yeonju Kim$^{1}$ \quad Bella Godiva$^{1}$ \quad Se Jin Park$^{2\dagger}$ \quad Yong Man Ro$^{1\dagger}$ \\
$^{1}$Integrated Vision and Language Lab, KAIST, South Korea \quad
$^{2}$Kyung Hee University, South Korea \\
\texttt{\{yeonju7.kim, bellagodiva, ymro\}@kaist.ac.kr}, \quad \texttt{sejinpk@khu.ac.kr} \\
$^{\dagger}$Corresponding authors.
}

\iclrfinalcopy 
\begin{document}

\maketitle
\lhead{Preprint}

\begin{abstract}
Full-duplex speech models listen and speak at the same time, which makes
them promising for natural, open-ended conversation. But can they sustain
a coherent conversation for tens of minutes? Because they generate a token every frame, even while listening, they
exhaust their context windows within minutes. Yet existing benchmarks
evaluate only single turns or a few conversational rounds, too short to
expose the long-horizon problem. We introduce the Full-Duplex
Long-Horizon Benchmark (FD-LHB), which tests five recent full-duplex models on long replayed
conversations and in live conversation with an LLM partner. Every evaluated model eventually either degrades or exhausts its context. Models that keep the full history often degrade before their context runs out, while sliding-window models still fall into \emph{repetition}, \emph{silence}, and \emph{gibberish}. For sliding-window models, we identify two contributing mechanisms. Gibberish and silence are linked to Initialization-Anchor Eviction,
where the sliding window discards the first KV entry. Repetition is also reinforced by the model's own recent outputs. Guided by this diagnosis, we apply two simple fixes:
Initialization-Anchor Pinning (IAP) keeps that first entry, and
Duplex Repetition Penalty (DRP) discourages echoing while guarding
against prolonged silence. Together, they support the diagnosis and, without retraining, substantially reduce conversational collapse in replays of up to an hour while running in real time.
\end{abstract}

\section{Introduction}
\label{sec:intro}

\definecolor{fdcoherent}{RGB}{46,125,50}
\definecolor{fdcoherentbg}{RGB}{238,248,239}
\definecolor{fdrepetition}{RGB}{217,119,6}
\definecolor{fdrepetitionbg}{RGB}{255,247,232}
\definecolor{fdsilence}{RGB}{40,120,181}
\definecolor{fdsilencebg}{RGB}{237,246,252}
\definecolor{fdgibberish}{RGB}{178,58,138}
\definecolor{fdgibberishbg}{RGB}{252,239,248}
\definecolor{fdneutral}{RGB}{85,85,85}
\definecolor{fdneutralbg}{RGB}{245,245,245}
\newcommand{\fdexamplebox}[4]{%
    \noindent\fcolorbox{#1}{#2}{%
    \parbox{\dimexpr\linewidth-2\fboxsep-2\fboxrule\relax}{%
    \scriptsize\textbf{#3}\enspace #4}}\par\vspace{1.5pt}}
\begin{wrapfigure}{r}{0.32\linewidth}
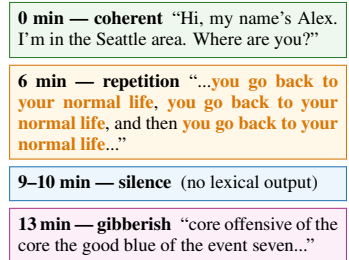

    \vspace{-1.2em}
    \centering
    \fdexamplebox{fdcoherent}{fdcoherentbg}{0 min --- coherent}
    {``Hi, my name's Alex. I'm in the Seattle area. Where are you?''}
    \fdexamplebox{fdrepetition}{fdrepetitionbg}{6 min --- repetition}
    {``...\textcolor{fdrepetition}{\textbf{you go back to your normal life}},
    \textcolor{fdrepetition}{\textbf{you go back to your normal life}}, and then
    \textcolor{fdrepetition}{\textbf{you go back to your normal life}}...''}
    \fdexamplebox{fdsilence}{fdsilencebg}{9--10 min --- silence}
    {(no lexical output)}
    \fdexamplebox{fdgibberish}{fdgibberishbg}{13 min --- gibberish}
    {``core offensive of the core the good blue of the event seven...''}
    \vspace{-0.3em}
    \captionsetup{font=small, skip=3pt, justification=raggedright,
                  singlelinecheck=false}
    \caption{\textbf{Example of long-conversation collapse.} It shows progression of coherent speech, repetition, silence, and gibberish, with timestamps aligned to the  timeline in Figure 2a.}
    \label{fig:representative_failure_progression}
\end{wrapfigure}

Human conversation can continue for tens of minutes or longer, as speakers move between long turns, silence, interruptions, backchannels (e.g., ``uh-huh'' or ``yeah''), and overlapping speech~\citep{levinson2015timing}. Full-duplex speech models~\citep{defossez2024moshi,roy2026personaplex,kim2026raon,fang2026bayling} aim to support this form of interaction by listening and speaking concurrently. Maintaining this behavior over long conversations remains difficult because, unlike text LLMs or half-duplex systems, full-duplex models continuously generate tokens frame by frame, even while listening. This rapidly fills the context with silence and recurring backchannels, while each generated output also becomes part of the next input and can influence later generation.

Despite these challenges, long-horizon behavior in full-duplex models remains poorly understood. Existing full-duplex benchmarks~\citep{lin2025full,
lin2026full15,lin2026full2,peng2025fd,arora2025talking,he2026mtr} evaluate
capabilities such as turn-taking, interruption handling, pauses,
backchanneling, and overlapping speech, within a single turn or a few
conversational rounds, not over many minutes. Existing models manage conversation history in one of two ways.
Growing-context models~\citep{kim2026raon,fang2026bayling} keep the full
interaction, so their context grows until it reaches a context limit.
Sliding-window models~\citep{defossez2024moshi,roy2026personaplex} keep only
the most recent frames, discarding the oldest as new ones arrive, so they can
run indefinitely with fixed-size cache. Whether either design is sufficient for
long-form full-duplex interaction, and how the models fail when it is not,
has not been systematically investigated.

\begin{figure}[t]
    \centering
    \includegraphics[width=\linewidth]{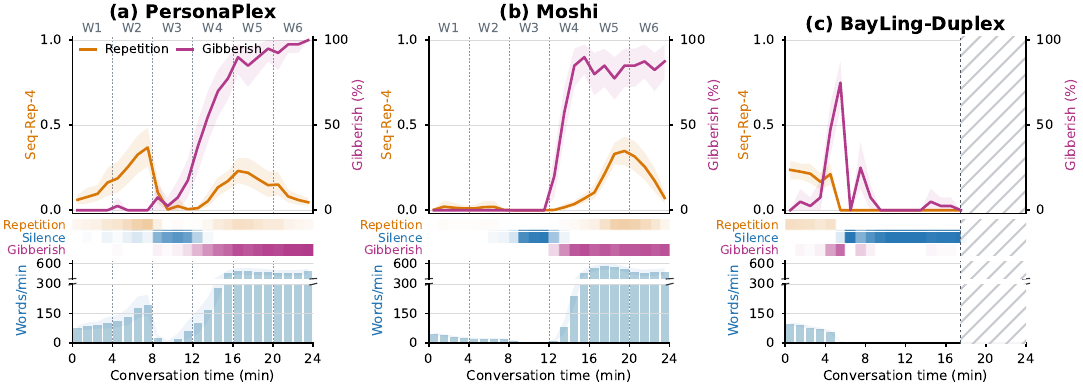}
\caption{\textbf{How three models degrade over a long conversation.} Top: repetition and gibberish rate; middle: occurrence of each failure mode; bottom: output words per minute. Sliding-window models (a) PersonaPlex and (b) Moshi exhibit an initial period of coherent speech, followed by silence and eventual gibberish, with repetition emerging at different times. The growing-context model (c) BayLing-Duplex develops gibberish and silence before reaching its context limit.}
    \label{fig:population_failure_patterns}
    \vspace*{-1.7em}
\end{figure}


To investigate this problem, we introduce the \textbf{Full-Duplex Long-Horizon Benchmark (FD-LHB)}, a benchmark designed specifically to evaluate full-duplex models over extended interaction. FD-LHB combines controlled replay of CANDOR conversations~\citep{reece2023candor} with adaptive interaction against an LLM-based spoken partner. Using FD-LHB, we find that every evaluated model eventually either degrades, through repetition, silence, or gibberish (\cref{fig:representative_failure_progression}), or exhausts its context: sliding-window models collapse even though their memory never runs out (\cref{fig:population_failure_patterns}a,b), and growing-context models can collapse before reaching their context limit (\cref{fig:population_failure_patterns}c).

We then ask why these collapses occur, focusing on the two
sliding-window models, and identify two contributing mechanisms
(\cref{sec:diagnosis}). Silence and gibberish are linked to
\textbf{Initialization-Anchor Eviction}: the sliding window eventually
discards the first KV entry computed without prior cached context.
We hypothesize that this dependence reflects a mismatch between
empty-cache training and continual inference. We also identify
\textbf{Self-Reinforcing Repetition}: a word the model has recently said
becomes more likely to be said again.

Finally, these mechanisms motivate two simple fixes that need no
retraining. \emph{Initialization-Anchor Pinning (IAP)} keeps the first KV
entry in the cache while the rest of the window slides as usual,
and \emph{Duplex Repetition Penalty (DRP)} discourages recently used words without making the model less likely to speak. Together, they substantially
reduce repetition, silence, and gibberish in both
sliding-window models. In addition, the combined method still runs in real
time and is faster than a StreamingLLM-style baseline~\citep{xiao2024efficient}
while matching its cache stability. We do not claim that they solve
long-horizon full-duplex modeling; rather, our results show that simple fixes
targeting the two identified causes prevent much of the collapse.

Our contributions are as follows:
\begin{itemize}
\item \textbf{FD-LHB: benchmark for long-horizon full-duplex interaction.} FD-LHB measures when and how conversational failures emerge through two tracks:  40 controlled replays evaluated up to 60 minutes (FD-LHB-CANDOR) and 40 adaptive sessions up to 20 minutes (FD-LHB-LLM).
\item \textbf{Systematic characterization and diagnosis of collapse.} Across five recent full-duplex models using sliding-window or growing-context, we characterize the onset and progression of three failure modes: repetition, prolonged silence, and gibberish. Experiments on two sliding-window models identify Initialization-Anchor Eviction and Self-Reinforcing Repetition as mechanisms underlying the collapse.
\item \textbf{Validation through training-free interventions.} Simple, targeted fixes reduce gibberish from 100\% to 0\%. Repetition falls from 65\% to 0\% in Moshi and 90\% to 2.5\% in PersonaPlex; prolonged silence falls from 100\% to 7.5\% and 85\% to 47.5\%, respectively. These interventions require no retraining and can be incorporated into compatible baselines at inference time. 
\end{itemize}

\section{Related Work}
\label{sec:related_works}
\subsection{Full-Duplex Speech Generation and Streaming Interaction}

Models such as Moshi~\citep{defossez2024moshi},
PersonaPlex~\citep{roy2026personaplex}, Raon-SpeechChat~\citep{kim2026raon},
BayLing-Duplex~\citep{fang2026bayling}, and Lychee-FD~\citep{liu2026lychee}
support low-latency full-duplex
interaction. Existing benchmarks~\citep{lin2025full,lin2026full15,lin2026full2,peng2025fd,
arora2025talking,he2026mtr,wang2026humdial,ray2026tau} evaluate turn-taking, overlap, interruption handling, multi-round quality, and task completion, but do not examine their stability
over extended conversations. We address this gap by evaluating repetition, silence, and gibberish during continuous generation lasting tens of minutes.

\subsection{Long-Horizon Stability and KV-Cache Management}

Autoregressive generation can suffer from degeneration in
text~\citep{holtzman2019curious}, video~\citep{ge2022long}, and
speech~\citep{song2025ella,chen2024vall}. Scheduled Sampling~\citep{bengio2015scheduled} and Professor
Forcing~\citep{lamb2016professor} address the mismatch between training and generation through modified training procedures. We instead investigate instability during continual
full-duplex inference, where generated outputs can reinforce later
repetition~\citep{xu2022learning,zhu2023penalty}. DRP targets this feedback by
modifying token selection without retraining.

Beyond repetition, continual inference raises the question of
how conversation history should be retained. Under a fixed memory budget,
StreamingLLM~\citep{xiao2024efficient} preserves initial attention sinks
together with recent tokens, as do LM-Infinite~\citep{han2024lminfinite} and InfLLM~\citep{xiao2024infllm}; H2O~\citep{zhang2023h2o} retains tokens with high
accumulated attention; and SnapKV~\citep{li2024snapkv} selects prompt positions
using attention patterns from an observation window. Building on these studies, we examine the role of the first KV entry in the long-horizon stability of full-duplex speech models. StreamingLLM is the closest to IAP because it also preserves entries
from the beginning of the sequence. In Moshi and PersonaPlex, the first KV entry alone prevents
gibberish in all tested runs, whereas the tested
later entries do not do so consistently (\cref{sec:diagnosis_firstentry}).
We hypothesize that this dependence reflects initialization-anchor eviction: training chunks
always begin with this entry, but long inference evicts it
(\cref{sec:chunk_rollout_mismatch}). IAP keeps only this
entry and updates only
its position index, which is cheaper than re-indexing the whole cache as
StreamingLLM does (\cref{sec:efficiency}).

\section{Full-Duplex Long-Horizon Benchmark}
\label{sec:fd_lhb}

\subsection{Evaluation Protocol}
\label{sec:fd_lhb_protocol}
Existing benchmarks do not test whether these models remain stable over sustained interactions
lasting tens of minutes. To address this gap, we introduce the
Full-Duplex Long-Horizon Benchmark (FD-LHB).

FD-LHB has two complementary tracks: controlled replay
(\textbf{FD-LHB-CANDOR}) and adaptive interaction (\textbf{FD-LHB-LLM}). Replay (FD-LHB-CANDOR) gives every model the same recorded human speech,
enabling controlled comparison, but it is non-interactive: the input doesn't respond to the model's output. Adaptive (FD-LHB-LLM), by contrast, is interactive, with an LLM partner whose synthesized speech reacts to the model's behavior like a real conversation, but this means the input differs across runs, making controlled comparison harder. Because each track compensates for the other's weakness, the two are complementary. 


\paragraph{FD-LHB-CANDOR.}
FD-LHB-CANDOR is built from CANDOR~\citep{reece2023candor}, a corpus of long, natural human conversations.
We remove the first three turns from each
conversation to exclude greetings and initial silence, rank the conversations
by their remaining duration, and select the longest 40. The resulting streams
have minimum, mean, and maximum durations of 55.5, 67.4, and 92.6 minutes,
respectively. We evaluate each replay for up to 60 minutes, or until the available conversation ends. For each conversation, we select the speaker channel containing
more speech and use it as the fixed audio input to the evaluated model. Thus,
although the original CANDOR recordings are interactive human conversations,
FD-LHB-CANDOR replays the same fixed input to every model, enabling controlled
comparison.

\paragraph{FD-LHB-LLM.}

To complement the non-interactive fixed replay of FD-LHB-CANDOR, FD-LHB-LLM evaluates the model in a closed-loop conversation with an LLM-based spoken partner whose responses adapt to the evaluated model's output. At each turn, Gemini-2.5-Flash~\citep{comanici2025gemini}
generates the partner's next text response from the recent dialogue history,
and a local \texttt{facebook/mms-tts-eng} VITS model~\citep{pratap2023mms}
converts it to 16-kHz speech for the evaluated model. The evaluated model's response is then returned to Gemini, which generates the next partner turn. Each session begins with one of ten everyday-life topics, such as hobbies, travel, family, or work, and the partner responds adaptively to the evaluated model rather than following a fixed script. We run 40 sessions per dialogue model, each for up to 20 minutes or until the model reaches its context capacity.

\subsection{Failure Modes and Onset Metrics}
\label{sec:failure_detection}

As shown in \cref{fig:representative_failure_progression} and
\cref{fig:population_failure_patterns}, we observe three main forms of
long-horizon collapse: \textbf{repetition}, \textbf{silence}, and
\textbf{gibberish}.
To measure when each form first appears, we define its \emph{onset time}.
Repetition is evaluated over 32-word windows, while silence and gibberish are
evaluated in one-minute bins.
For each form, the onset is the start of the first window or bin that meets
its criterion.


\paragraph{Repetition.}
We measure repetition with word-level Seq-Rep-4~\citep{xu2022learning},
which captures repeated four-word patterns in the model's recent output.
We use this metric because it detects recurring phrase-level patterns.
We compute it over 32-word windows and define a window as repetitive when
Seq-Rep-4 $\geq 0.30$; examples around this threshold are shown in
\cref{tab:seqrep_examples}.
The repetition onset is the start of the first of two consecutive repetitive
windows, so a single repeated phrase is not counted.

\paragraph{Silence.}
We measure silence using the number of spoken words in each bin.
A bin is considered silent when it contains fewer than three words.
To exclude short conversational pauses, we define prolonged-silence onset as the first bin of two consecutive silent bins.

\paragraph{Gibberish.}
We characterize gibberish using two measures: a Gemini-2.5-Flash judge and
predictive entropy, computed from the model's distribution over its next text
token. Gemini classifies each one-minute speaking bin as either gibberish or
non-gibberish, while predictive entropy measures the model's uncertainty.
We evaluate bins containing at least three words and define the gibberish onset
as the first bin classified as gibberish by Gemini. A blinded listening study with eight raters shows that the judge never
calls normal speech gibberish but sometimes misses real gibberish
(\cref{sec:human_gibberish_validation}), so the reported gibberish rates
may be slightly lower than the true rates. We do not use external-LM perplexity; \cref{sec:ppl_coherence_audit}
explains why.

\begin{wrapfigure}{R}{0.5\textwidth}
    \centering
    \includegraphics[width=\linewidth]{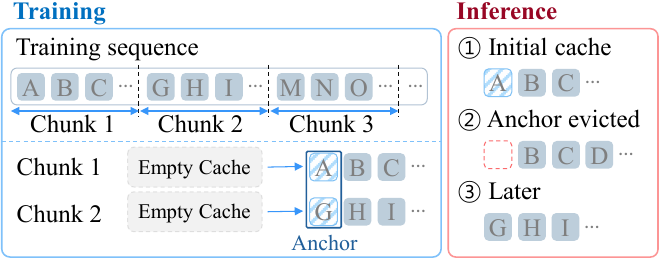}
    \caption{\textbf{Initialization-Anchor Eviction.}
    When training starts each chunk from an empty cache, every chunk begins
with a fresh initialization anchor. In sliding-window inference, the cache
is never reset, so the initial anchor is eventually evicted.}
    \label{fig:chunk_rollout_mismatch}
\end{wrapfigure}

\section{Diagnosing Long-Conversation Collapse}
\label{sec:diagnosis}
Based on FD-LHB (\cref{sec:fd_lhb}), we next investigate the mechanisms behind the three forms of collapse observed in \cref{fig:population_failure_patterns}:
repetition, silence, and gibberish.

\subsection{Initialization-Anchor Eviction}
\label{sec:chunk_rollout_mismatch}
The similar timing of silence and gibberish in Moshi~\citep{defossez2024moshi} and PersonaPlex~\citep{roy2026personaplex} shown in \cref{fig:population_failure_patterns} motivates us to examine their shared cache-management strategy.
Both models use a sliding window that discards the oldest cached entries as new ones arrive.
As shown in \cref{fig:chunk_rollout_mismatch}, training splits each sample
into chunks that start from an empty cache, so every chunk begins with an
\emph{initialization anchor}: the first KV entry, computed without any
earlier cached state. We confirmed this setup in the official Moshi
fine-tuning code; whether pretraining and PersonaPlex training follow it is
not public (\cref{app:anchor_analysis}). At inference, however, the cache is
never reset. Once it is full, the oldest entries are evicted, starting with
the initialization anchor. Training therefore always contains an anchor,
whereas long inference eventually does not, creating a training--inference
mismatch. We hypothesize that this initialization-anchor eviction causes the
collapse and test this hypothesis in \cref{sec:collapse_anchor_turnover,sec:diagnosis_firstentry}.

\subsubsection{Collapse Timing Follows Cache Turnover}
\label{sec:collapse_anchor_turnover}

\begin{figure}[t]
    \centering
    \includegraphics[width=\linewidth]{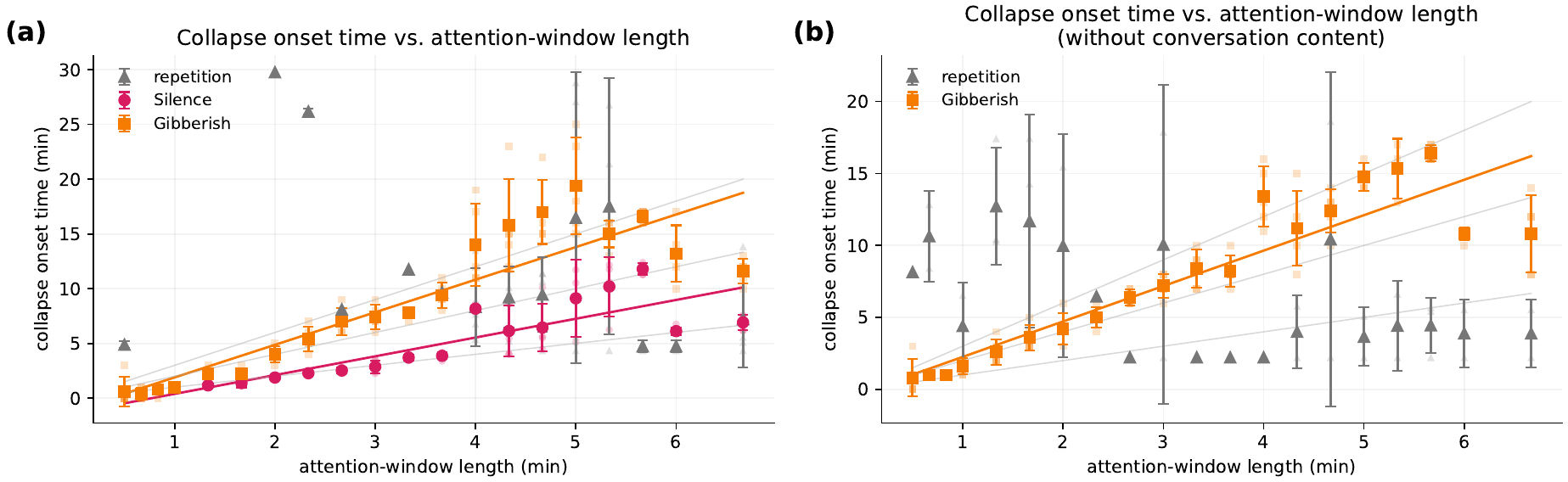}
    \caption{
    \textbf{Silence and gibberish track cache turnover.}
    Collapse onset time versus attention-window length $W$.
    (a) Conversational input. (b) Silence-only input.
    Silence and gibberish begin later as $W$ grows, roughly in proportion to $W$.}
   \label{fig:collapse_window_scaling}
\end{figure}

Based on the initialization-anchor eviction hypothesis
(\cref{sec:chunk_rollout_mismatch}), we vary the
attention-window length $W$ and measure when each type of collapse begins
(\cref{fig:collapse_window_scaling}). Results from Moshi and PersonaPlex are
combined. 

\Cref{fig:collapse_window_scaling}a shows the results with conversational audio
input. Silence onset grows at
$(1.63{\pm}0.49)W$ and gibberish onset at $(2.70{\pm}0.20)W$. Repetition, in
contrast, varies widely across runs and shows no clear relation to $W$. To separate cache turnover from conversational content, \cref{fig:collapse_window_scaling}b repeats the sweep with silence-only input.
Gibberish still appears and grows at $(2.54{\pm}0.19)W$, whereas repetition
does not start later as $W$ grows. These results show that silence and gibberish onsets scale roughly with
$W$, with gibberish following this pattern even in the absence of conversational content, linking these failures to cache turnover, while repetition shows no
consistent relationship. We investigate repetition further in
\cref{sec:self_reinforcing_repetition}.


\subsubsection{The first KV entry is sufficient}
\label{sec:diagnosis_firstentry}

\begin{figure*}[t!]
    \centering
    \includegraphics[width=0.96\textwidth]{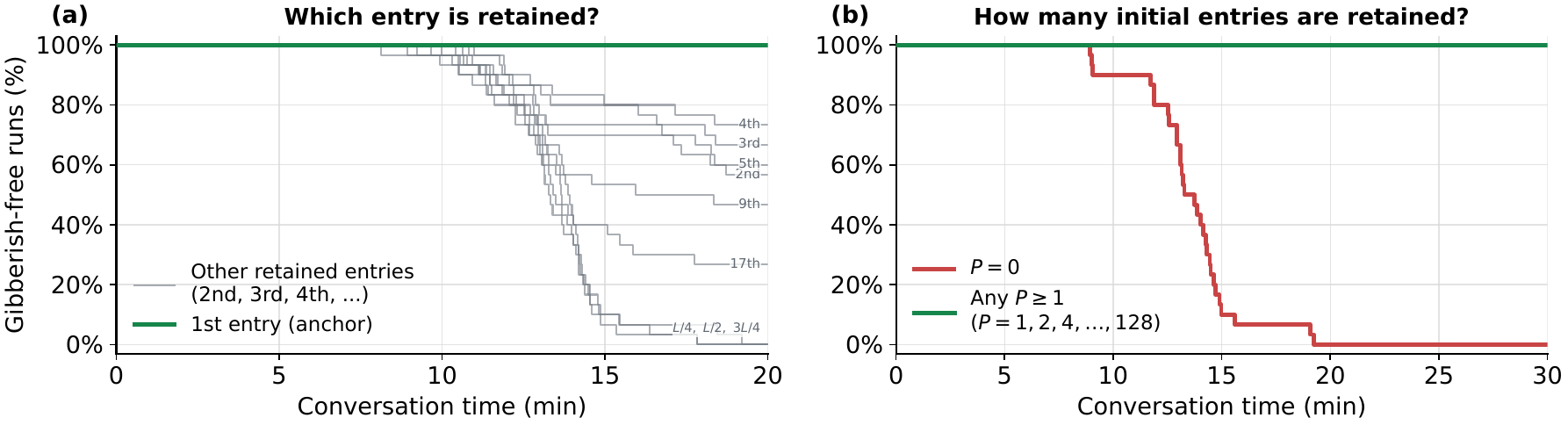}
    \caption{\textbf{Retaining the first entry is sufficient.} Percentage of runs without gibberish over time.
    (a) Which single early entry is kept.
    (b) Number of initial entries kept, $P$.}
    \label{fig:anchor_identity_count}
\end{figure*}

Based on the initialization-anchor eviction hypothesis
(\cref{sec:chunk_rollout_mismatch}), we investigate \emph{which} cache entry and
\emph{how many} entries must be retained to prevent gibberish. In both
experiments below, we measure the percentage of runs that do not produce
gibberish over time (\cref{fig:anchor_identity_count}). Results from Moshi and
PersonaPlex are combined.

In \cref{fig:anchor_identity_count}a, we keep exactly one early entry in the
cache and vary only which one it is: the 1st, 2nd, 3rd, 4th, 5th, 9th, or
17th entry, or the entry at $L/4$, $L/2$, or $3L/4$, where $L$ is the window
length. Only the entry itself changes; its place in the cache and its position
index stay the same. Surprisingly, only the first entry prevents gibberish in all runs;
every other choice fails in at least one run.
In \cref{fig:anchor_identity_count}b, we keep the first $P$
consecutive entries, with $P \in \{0, 1, 2, 4, 8, 16, 32, 64, 128\}$. With
$P{=}0$, every run reaches gibberish. With any $P{\geq}1$, no run does, and
the curves overlap exactly. Together, these results show that the first entry alone prevents gibberish
in all tested runs, and keeping more initial entries gives no further
improvement.

\begin{figure*}[t!]
    \centering
    \includegraphics[width=\textwidth]{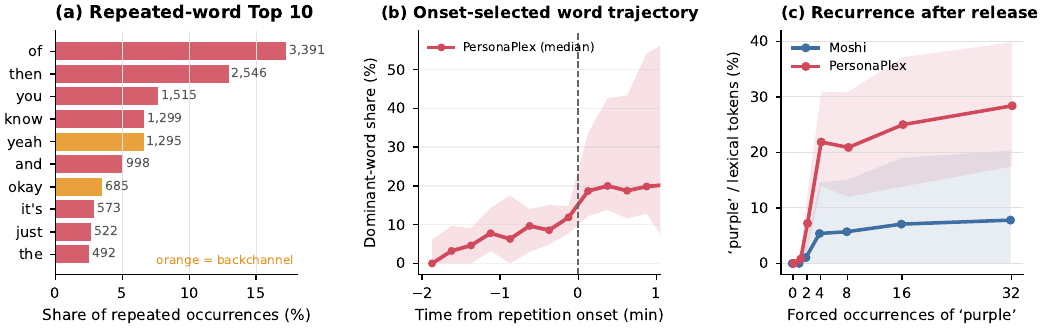}
\caption{\textbf{Recent outputs reinforce repetition.}
(a) Top 10 repeated words during repetition periods; orange marks backchannels.
(b) Share of each PersonaPlex run's dominant word around repetition onset; seven diagnostic runs without a sustained onset are
excluded.
(c) Share of \emph{purple} in the 20\,s after inserting it $d$ times.}
    \label{fig:self_reinforcing_repetition}
\end{figure*}

\subsection{Recent Outputs Reinforce Repetition}
\label{sec:self_reinforcing_repetition}

We next examine whether repetition in full-duplex speech is driven by
self-reinforcement. In text generation, repeated content in the context makes
the same content more likely, so a loop can reinforce
itself~\citep{xu2022learning,zhu2023penalty}. Based on this, we hypothesize
that recent outputs similarly reinforce repetition in full-duplex speech, and
test this hypothesis in \cref{fig:self_reinforcing_repetition}.
First, the most repeated words during repetition periods are short, frequent
words such as \emph{of}, \emph{then}, and \emph{you}, and backchannels such as
\emph{yeah} and \emph{okay} (\cref{fig:self_reinforcing_repetition}a).
Second, for each PersonaPlex run we select the word that dominates during
repetition periods and track how frequently it occurs. It starts rising before
onset and keeps rising afterward
(\cref{fig:self_reinforcing_repetition}b). This suggests that a repetition loop
does not appear suddenly, but grows from a word the model has already been
saying.
Third, we insert \emph{purple}, a word absent from the
previous context, and count how often the model repeats it over the next 20\,s.
Recurrence increases with more insertions in both models, most strongly in
PersonaPlex (\cref{fig:self_reinforcing_repetition}c).
Together, these results show that recent exposure makes a word recur.

\section{Testing the Diagnosis with Training-Free Interventions}
\label{sec:methodology}

The diagnosis yields two testable predictions: retaining the initialization anchor should reduce failures associated with cache eviction, and limiting reinforcement from recent words should reduce repetition. We test these predictions with Initialization-Anchor Pinning (IAP) and Duplex Repetition Penalty (DRP), respectively. Both are training-free methods that operate at inference time and can be applied to existing models.

\subsection{Initialization-Anchor Pinning (IAP)}
\label{sec:iap}

\Cref{sec:diagnosis_firstentry} showed that keeping the first KV entry is
enough to prevent gibberish, while keeping more initial entries adds
nothing. IAP therefore keeps this initialization anchor instead of letting the
sliding window evict it. Until the cache is full, nothing changes. Once it is full, the model normally
drops the oldest entry each time a new one arrives. IAP does the same, except
that it never drops the anchor: when a new entry arrives, it removes the
oldest entry after the anchor instead. This is done in every layer. As in StreamingLLM~\citep{xiao2024efficient},
the anchor's position index stays next to the rest of the window so
that position indices remain continuous. StreamingLLM, however,
re-indexes the entire cache as the window slides. IAP instead
simply increases the anchor's index by one each time the window slides. Only
this single entry changes, which keeps IAP cheap (\cref{sec:efficiency}).

\subsection{Duplex Repetition Penalty (DRP)}
\label{sec:drp}

DRP reduces repeated words without letting the model fall silent instead.
It changes only the assistant's text-token logits; the audio streams are
unchanged.

DRP penalizes any word that appeared in the model's last $N{=}512$ lexical tokens, using the standard repetition penalty of CTRL~\citep{keskar2019ctrl}:
a positive logit is divided by $\rho$ and a negative one is multiplied
by it. A word is penalized based only on whether it appeared recently, not on
how often. \texttt{PAD} and \texttt{EPAD}, which the model uses for listening
and turn-taking~\citep{roy2026personaplex,defossez2024moshi}, are excluded
from this penalty. Penalizing words lowers their total probability and makes
\texttt{PAD} and \texttt{EPAD} relatively more likely. To undo this, we raise all word logits by the same amount, so that the total probability of words stays the same as before the penalty. This removes the relative boost that the penalty gives to PAD and EPAD.

With this penalty alone, repetition decreases, but \texttt{PAD} can still
dominate for long periods, leaving the model silent. To address this, when
the model has been silent for a long time and the user is quiet, we
gradually lower the \texttt{PAD} logit. We do not force any token; the model
still samples its next token as usual (\cref{app:drp}).

\section{Results and Analysis}
\label{sec:results}

\subsection{Experimental Setup}
\label{sec:setup}

We evaluate five full-duplex models using two context-management strategies:
sliding-window and growing-context.
Moshi~\citep{defossez2024moshi} and PersonaPlex~\citep{roy2026personaplex}
are sliding-window models with a 3,000-frame window at 12.5\,Hz, so their
window first fills after four minutes. Raon-SpeechChat~\citep{kim2026raon},
BayLing-Duplex~\citep{fang2026bayling}, and Lychee-FD~\citep{liu2026lychee}
are growing-context models that run until their context is full, after about
5, 17.5, and 10.9 minutes, respectively. Reaching context capacity ends the
run; this limits sustained operation but is not counted as a collapse event.
In \cref{fig:candor_failure_free}, such runs are right-censored: they count as
failure-free until they end, and the curves are estimated with the
Kaplan--Meier method~\citep{kaplan1958nonparametric}. We apply IAP+DRP to the two sliding-window models,
since anchor eviction occurs only when a sliding window discards old entries.
Generation settings for each model are given in \cref{tab:repro_settings}.

\subsection{Main Results}
\label{sec:main_results}

\begin{figure}[t!]
    \centering
    \includegraphics[width=\linewidth]{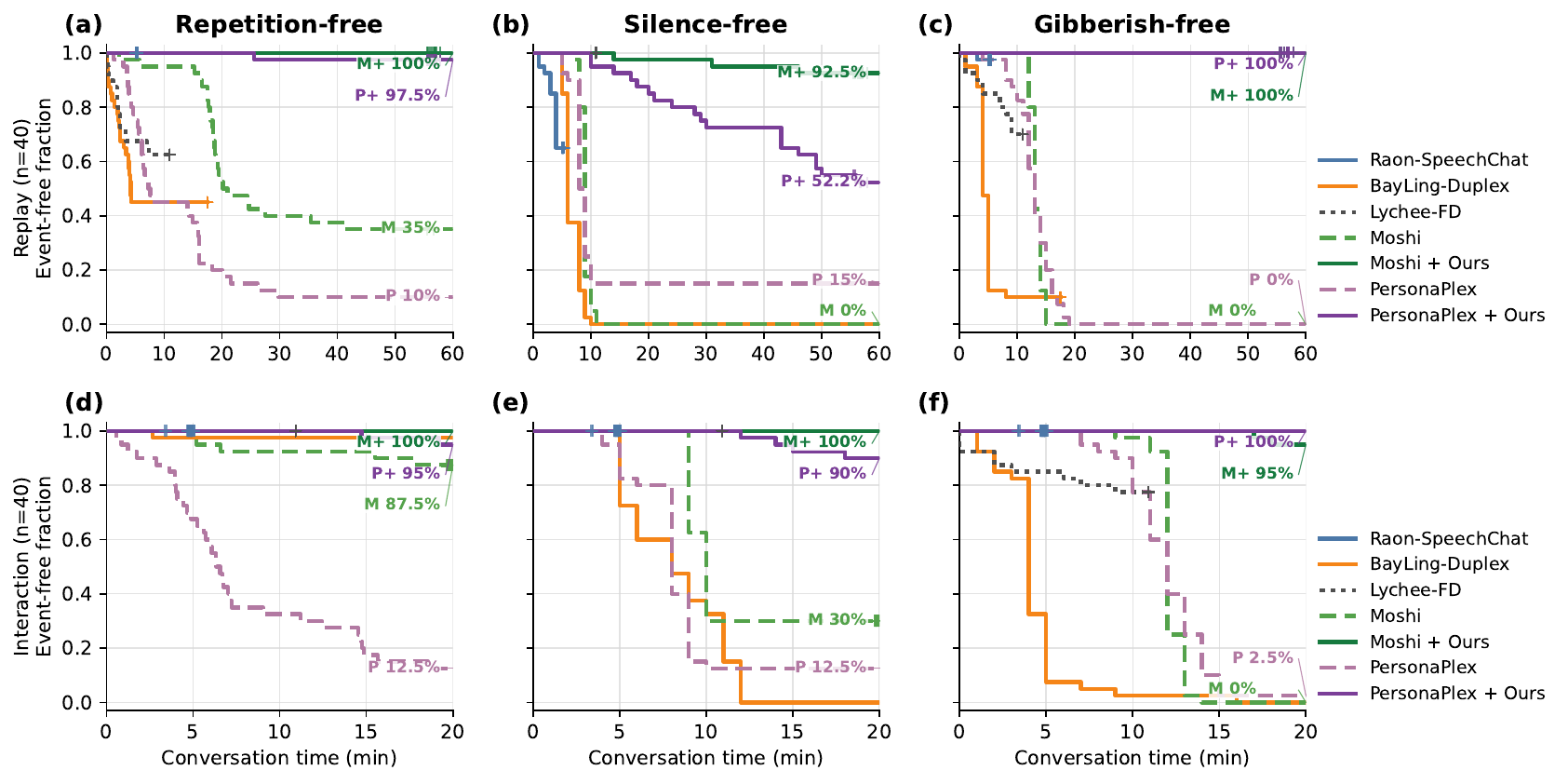}
\caption{\textbf{Repetition, silence, and gibberish across replay and adaptive interaction.}
Top: FD-LHB-CANDOR; bottom: FD-LHB-LLM. Curves show the percentage of runs remaining free of repetition (a, d), prolonged silence (b, e), or gibberish (c, f) over time; higher is better. For growing-context
models, a plus marks where the model reaches its context
capacity. IAP+DRP largely suppresses repetition and gibberish in both models and reduces prolonged silence.}
    \label{fig:candor_failure_free}
\end{figure}

\Cref{fig:candor_failure_free} shows how long each system stays free of
each collapse on FD-LHB-CANDOR (top) and FD-LHB-LLM (bottom); ``Ours''
denotes IAP+DRP. Observation lengths differ across
models, so we compare methods within each model rather than across models.
All percentages below are computed over 40 runs per system.

\paragraph{Every baseline fails, in one of two ways.}
On FD-LHB-CANDOR (\cref{fig:candor_failure_free}, top), growing-context models
run out of context or collapse first: Raon-SpeechChat reaches capacity after
about five minutes, BayLing-Duplex shows repetition, silence, and gibberish in
55\%, 100\%, and 90\% of runs within 17.5 minutes, and Lychee-FD shows
repetition and gibberish in 37.5\% and 30\% within 10.9 minutes.
Sliding-window models never exhaust their context but still collapse: over 60
minutes, Moshi and PersonaPlex show the three failures in 65/100/100\% and
90/85/100\% of runs. FD-LHB-LLM shows the same pattern
(\cref{fig:candor_failure_free}, bottom). Thus, none of the evaluated models
sustains stable conversation over tens of minutes, regardless of how it manages
its context.

\paragraph{IAP+DRP reduces collapse.}
We apply IAP+DRP to the two sliding-window models. On FD-LHB-CANDOR, the
repetition, silence, and gibberish rates drop to 0\%, 7.5\%, and 0\% for
Moshi, and 2.5\%, 47.5\%, and 0\% for PersonaPlex. The PersonaPlex+Ours
silence in \cref{fig:candor_failure_free}b is mostly temporary: 94.7\% of the
affected runs resume speaking within a minute after the two-minute silence.
On FD-LHB-LLM, where the partner responds to the model, the baselines also fail
frequently, most often through gibberish (97.5--100\% of sessions) and
prolonged silence (70--87.5\%). With IAP+DRP, each failure occurs in no more than 10\% of sessions for both
models.
\Cref{tab:main_results} lists all counts and onset times. Thus, without
retraining, IAP+DRP substantially reduces all three forms of collapse in both
sliding-window models, in both controlled replay and adaptive interaction.

\begin{wrapfigure}[15]{r}{0.60\linewidth}
    \vspace{-0.8em}
    \centering
    \includegraphics[width=0.98\linewidth]{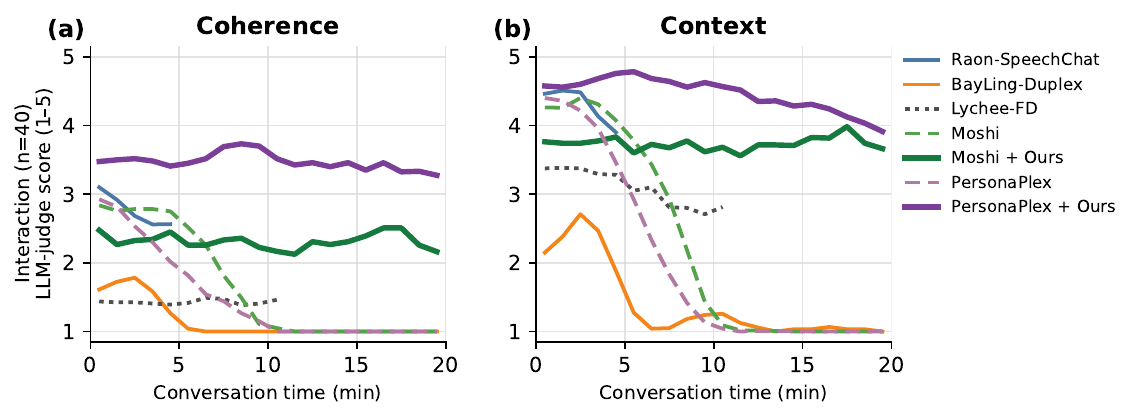}
\caption{\textbf{LLM-judge scores over time.} Mean Gemini-2.5-Flash ratings on FD-LHB-LLM for (a) coherence and (b) context to the partner’s preceding speech.}
    \label{fig:interaction_llm_judge}
    \vspace{-0.5em}
\end{wrapfigure}

\paragraph{IAP+DRP also preserves response quality.}
Avoiding collapse does not guarantee the quality of what the model says.
We therefore also score the content of each one-minute speaking bin on
FD-LHB-LLM, using Gemini-2.5-Flash as a judge (\cref{fig:interaction_llm_judge}). It rates two aspects on a
1--5 scale: \emph{Coherence}, whether the model's speech is fluent and makes
sense, and \emph{Context}, whether it responds appropriately to what the
partner said. Silent bins receive a score of 1.
Both baselines fall to the lowest score within about ten minutes, whereas
IAP+DRP keeps both scores steady for the full session.
\Cref{app:trajectories} reports the same scores for FD-LHB-CANDOR, and
\cref{app:components} for the ablation. Thus, IAP+DRP avoids collapse while keeping the model's speech coherent and
responsive.

\FloatBarrier

\subsection{Ablation}
\label{sec:ablation}

\begin{figure}[t!]
    \centering
    \includegraphics[width=\linewidth]{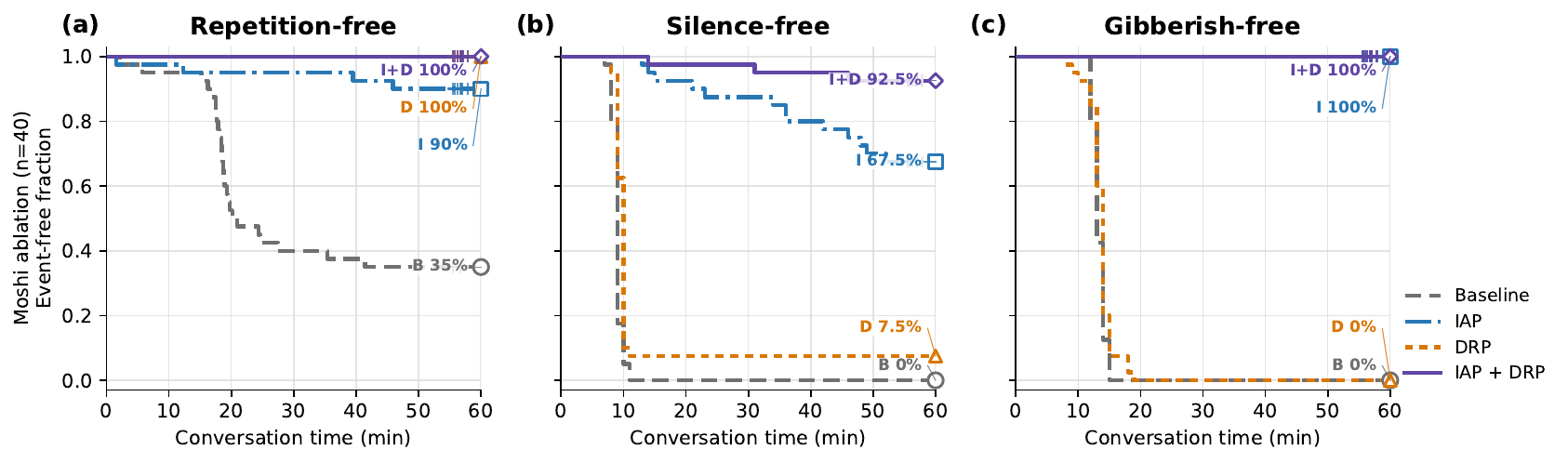}
    \caption{\textbf{Ablation of cache retention and repetition control.} Curves show the proportion remaining free of (a) repetition, (b) prolonged silence, and (c) gibberish. Endpoint labels identify Baseline (B), IAP (I), DRP (D), and IAP+DRP (I+D).}
    \label{fig:moshi_failure_free_ablation}
\end{figure}

To investigate the effect of each component, we apply IAP and DRP separately
and together to Moshi on the FD-LHB-CANDOR replays
(\cref{fig:moshi_failure_free_ablation}).
For repetition in \cref{fig:moshi_failure_free_ablation}a, DRP keeps every
run repetition-free, and IAP also raises the repetition-free share from 35\%
to 90\%, suggesting that initialization-anchor eviction contributes to repetition in part.
For silence in \cref{fig:moshi_failure_free_ablation}b, IAP raises the
silence-free share from 0\% to 67.5\%, and adding DRP raises it further to
92.5\%.
For gibberish in \cref{fig:moshi_failure_free_ablation}c, IAP alone keeps
every run free of Gemini-defined gibberish, whereas DRP alone does not help, supporting the link between gibberish and
initialization-anchor eviction. Thus, the two components are complementary: IAP addresses gibberish and most
silence, DRP addresses repetition, and only their combination suppresses all
three failures.

\subsection{Real-Time Inference Cost}
\label{sec:efficiency}
\begin{wraptable}{r}{0.44\linewidth}
\vspace{-0.8em}
\centering
\scriptsize
\setlength{\tabcolsep}{3pt}
  \caption{\textbf{Real-time inference cost}
  Gibberish-free runs (out of 40) and real-time factor (RTF).}
\label{tab:computational-efficiency-main}
\begin{tabular}{@{}lcc@{}}
\toprule
Method & Gibberish-free $\uparrow$ & RTF $\downarrow$ \\
\midrule
Baseline & $0/40$ & $0.5957{\pm}0.0020$ \\
StreamingLLM-style & $40/40$ & $1.0323{\pm}0.0029$ \\
Ours & $40/40$ & $0.7525{\pm}0.0023$ \\
\bottomrule
\end{tabular}
\vspace{-0.8em}
\end{wraptable}
Full-duplex models must respond in real time, so any extra computation
matters. We therefore measure the cost of IAP and DRP
(\cref{tab:computational-efficiency-main}). The table reports two numbers
for PersonaPlex: how many of 40 runs stay free of gibberish, and the real-time factor (RTF). RTF is the time, in seconds, needed to process one second of audio. An RTF
below one means the model runs faster than real time. We measure RTF five times on an RTX A6000.

We compare three methods. Baseline uses the standard FIFO cache.
StreamingLLM-style~\citep{xiao2024efficient} keeps the first four entries and re-indexes all remaining entries at every step. It avoids gibberish in every
run, but its RTF is 1.03, so it cannot keep up in real time. Ours also
avoids gibberish in every run, but reaches an RTF of 0.75, because IAP
re-indexes only the first entry. In short, both methods prevent gibberish, but IAP is faster because it updates
one entry per step instead of all of them.
\Cref{sec:supp-computational-efficiency} reports the cost of each component.

\section{Conclusion}
\label{sec:conclusion}

We introduced FD-LHB to evaluate whether full-duplex speech models can
sustain long conversations, using controlled replay and adaptive
interaction. Across five models, sustained operation was limited by
behavioral degradation or context exhaustion.
Controlled experiments link silence and
gibberish to Initialization-Anchor Eviction, and repetition to
self-reinforcement from recent outputs.
Guided by these findings, we applied two simple interventions: IAP
retains the initialization anchor, while DRP discourages repetition
and guards against prolonged silence.
In replays of up to an hour, their combination reduces repetition,
silence, and gibberish, and it also improves stability in 
adaptive conversations with an LLM partner.
These gains validate our diagnosis. Both methods require neither
retraining nor additional memory, and they run in real time. Overall, we identify how and why full-duplex models collapse in long conversations and show that simple, cause-specific fixes can prevent much of this collapse.

\subsection*{AI use statement}
We used Gemini-2.5-Flash to generate partner utterances in FD-LHB-LLM and to
provide the automated judgments described in the evaluation protocol. We
used OpenAI Codex to assist with code development, debugging, and running
analysis scripts. We used Anthropic Claude and OpenAI Codex to help draft and
revise parts of the manuscript and to help find related work. The authors
reviewed all AI-assisted text, verified the code by running it and checking
the resulting logs and outputs, and take responsibility for the final
content, claims, and artifacts.

\section*{Ethics Statement}
Our listening studies (\cref{app:human_validation,app:human_tradeoff}) were minimal-risk perceptual
evaluations. Participants were adult, English-proficient crowdworkers
recruited through Prolific, who provided informed consent before
participating and could withdraw at any time. They listened to short
clips of model-generated speech and answered forced-choice questions;
no personally identifying information was collected beyond Prolific's
anonymized IDs. Participants were compensated at \pounds 9.00/hour,
in line with Prolific's recommended rate.
The benchmark uses the CANDOR corpus, whose participants consented
to recording and research use at collection time
\citep{reece2023candor}; we access it under its research-access terms
and release only a manifest of conversation IDs and crop boundaries,
not the audio.

\bibliography{iclr2027_conference}
\bibliographystyle{iclr2027_conference}

\appendix
\crefalias{section}{appendix}
\crefalias{subsection}{appendix}
\crefalias{subsubsection}{appendix}
\section*{Appendix Roadmap}
\addcontentsline{toc}{section}{Appendix Roadmap}
\Cref{tab:roadmap} points to the supporting evidence for each part of the
main text. \Cref{app:evaluation} validates the evaluation,
\cref{app:diagnostics} analyzes the two collapse mechanisms,
\cref{app:implementation} specifies IAP and DRP, \cref{app:ablations} reports
extended results, ablations, and computational cost, and
\cref{app:exploratory} collects exploratory observations and limitations.

\begin{table}[h]
\centering
\footnotesize
\setlength{\tabcolsep}{4pt}
\renewcommand{\arraystretch}{1.2}
\caption{Supporting evidence for each part of the main text.}
\label{tab:roadmap}
\begin{tabularx}{\linewidth}{@{}>{\raggedright\arraybackslash}p{0.33\linewidth}
  >{\raggedright\arraybackslash}p{0.17\linewidth}
  >{\raggedright\arraybackslash}X@{}}
\toprule
Main-text claim & Appendix & Evidence \\
\midrule
Gemini gibberish labels are reliable (\cref{sec:failure_detection})
  & \cref{app:human_validation}
  & Blinded human listening study \\
Text criteria reflect audible failures (\cref{sec:failure_detection})
  & \cref{app:audio_validation}
  & Loudness-based silence and ASR-based repetition \\
Only the first KV entry matters (\cref{sec:diagnosis_firstentry})
  & \cref{app:anchor_analysis}
  & Keep a single early KV entry or a prefix of entries; position-index and
    cache-history controls \\
Recent outputs reinforce repetition (\cref{sec:self_reinforcing_repetition})
  & \cref{app:repetition_analysis}
  & Insert a new word and measure whether the model repeats it \\
IAP and DRP (\cref{sec:methodology})
  & \cref{app:implementation}
  & Update rules, execution order, and final hyperparameters \\
IAP+DRP reduces collapse (\cref{sec:main_results})
  & \cref{app:main_results,app:trajectories,app:silence}
  & Onset times, significance tests, quality scores, and silence durations \\
Each component contributes (\cref{sec:ablation})
  & \cref{app:components,app:drp_ablation,app:alternative_controls,app:hyperparameters}
  & Component, control, and hyperparameter ablations \\
Effect of DRP on speech naturalness
  & \cref{app:human_tradeoff}
  & Blinded A/B listening study early, middle, and late in the conversation \\
IAP+DRP runs in real time (\cref{sec:efficiency})
  & \cref{app:efficiency}
  & Per-frame latency and real-time factor \\
Beyond full-duplex dialogue (exploratory)
  & \cref{app:half_duplex}
  & Additional speech models \\
\bottomrule
\end{tabularx}
\end{table}

The mechanism analyses in \cref{app:anchor_analysis} measure stability by
predictive entropy; all benchmark results use the Gemini gibberish label.

\section{Evaluation Setup and Validation}
\label{app:evaluation}
This section describes the benchmark inputs and model settings
(\cref{app:setup}), the metrics and judges (\cref{app:metrics}), a human
validation of the Gemini gibberish judge (\cref{app:human_validation}),
audio-side checks of the collapse metrics (\cref{app:audio_validation}), why
we use predictive entropy instead of perplexity
(\cref{app:metric_sensitivity}), the sensitivity of repetition and gibberish criteria
(\cref{app:threshold_sensitivity}), the Gemini prompts
(\cref{app:prompts}), and how to reproduce the results
(\cref{app:reproduction}). All later analyses use these definitions.

\subsection{Benchmark Inputs and Model Settings}
\label{app:setup}

\paragraph{Replay input.}
As described in \cref{sec:fd_lhb_protocol}, we remove the first three turns of each
CANDOR conversation, keep the 40 longest, and use the speaker channel with
more speech as input. Every model and condition receives exactly the same
16-kHz mono audio for a given conversation.

\paragraph{Interactive partner.}
The partner uses Gemini~2.5~Flash~\citep{comanici2025gemini} for text and a
\texttt{facebook/mms-tts-eng} VITS model~\citep{pratap2023mms} for 16-kHz
speech. It sees the last 16 turns and replies in one or two casual sentences
(temperature 0.9), asking natural follow-up questions and reacting naturally
when the model's speech is unclear or repetitive. Each session opens with one
of ten everyday topics: hobbies, pets, food, travel, entertainment, work,
music, family, exercise, or reading. Each model has 40 sessions, each lasting
up to 20 minutes or until the model reaches its context limit. The complete
partner prompt is given in \cref{app:prompts}.

\paragraph{Model prompts and sampling.}
PersonaPlex is given the persona ``You enjoy having a good conversation. Have
a warm, casual chat about everyday life, family, and pets. Your name is
Alex.'' Moshi receives no text persona. For each model, Baseline and Ours use
the same prompt, input, and sampling settings, so the two differ only in IAP
and DRP. Full settings are listed in \cref{tab:repro_settings}.

\paragraph{Compute.}
All experiments run on NVIDIA RTX A6000 GPUs with bfloat16 weights.

\subsection{Metrics, Judges, and Aggregation}
\label{app:metrics}
We measure three failure types (repetition, silence, and gibberish) and two
quality scores (Coherence and Context). Each failure is summarized by its
onset time, defined below.

\paragraph{Repetition.}
We compute Seq-Rep-4 over rolling 32-word windows:
\[
  \text{Seq-Rep-4} = 1 - \frac{\#\,\text{unique word 4-grams}}{\#\,\text{word 4-grams}}.
\]
Windows advance by 16 words and cover at most 30 seconds of audio. A window is
repetitive if Seq-Rep-4${}\geq0.30$. Repetition onset is the start of the first of two consecutive repetitive
windows, so a brief repeat that makes only one window repetitive is not
counted.
\Cref{tab:seqrep_examples} shows actual outputs at different scores, and
\cref{tab:repetition_sensitivity} reports results under other thresholds,
window sizes, and a single-window rule.

\begin{table}[!htbp]
\centering
\scriptsize
\setlength{\tabcolsep}{3pt}
\renewcommand{\arraystretch}{1.08}
\caption{\textbf{Example outputs at different Seq-Rep-4 levels.} Each row
shows the 32-word window from PersonaPlex replays whose score is closest to
the target level. Below 0.30, a phrase recurs at most twice; at 0.30, the
same phrase already recurs three times within the window. We therefore use
0.30 as the repetition threshold (highlighted). Dup.\ is the number of
repeated 4-grams among the 29 in the window.}
\label{tab:seqrep_examples}
\begin{tabularx}{\linewidth}{@{}ccc>{\raggedright\arraybackslash}X@{}}
\toprule
Target & Observed & Dup. & Decoded model output spanning 32 lexical words \\
\midrule
0.1 & 0.103 & 3/29 & ``computer at work, and they keep telling me, oh it' got to be a laptop. It' got to be a laptop. And, you know, I don' know that it actually does because'' \\
0.2 & 0.207 & 6/29 & ``that'd be good. But I know here, like, if you leave it with a bank, yeah, they charge you. Yeah, I think here, like, if you leave it with a bank, they'' \\
\rowcolor{gray!18}
0.3 & 0.310 & 9/29 & ``know. And \textbf{you can just kind of get away}. \textbf{You can just kind of get away from} the Midwest. \textbf{You can just kind of get away from} that, you know, because it'' \\
0.4 & 0.414 & 12/29 & ``education that people get, you know, the better, you know, the better people the country will be. And, you know, the better, you know, the better, you know, the better, you know'' \\
0.5 & 0.517 & 15/29 & ``it's just you know it's just you know it's just you know it's just you know it's just you it's just you know it's you know it justs you know it's just'' \\
0.6 & 0.586 & 17/29 & ``different group of people. And then the next day, you know it's, you know, it's, you know it's, you know, it's, you know, it's, you know, it's, you know, it's, you know'' \\
0.7 & 0.690 & 20/29 & ``it justs you know it justs you know it's just you know it's just you know it's just you know it's just you know it's just you know it's just you know'' \\
0.8 & 0.793 & 23/29 & ``moving. Okay, so anyway, so I moving. Okay, so anyway, so I moving. Okay, so anyway, so I moving. Okay, so anyway, so I moving. Okay, so anyway, so I moving. Okay'' \\
0.9 & 0.897 & 26/29 & ``Yeah she's spoiled. Yeah, she's spoiled. Yeah, she's spoiled. Yeah, she's spoiled. Yeah, she's spoiled. Yeah, she's spoiled. Yeah, she's spoiled. Yeah, she's spoiled. Yeah, she's spoiled. Yeah, she's spoiled. Yeah, she's'' \\
1.0 & 0.966 & 28/29 & ``Yeah. Yeah. Yeah. Yeah. Yeah. Yeah. Yeah. Yeah. Yeah. Yeah. Yeah. Yeah. Yeah. Yeah. Yeah. Yeah. Yeah. Yeah Yeah. Yeah Yeah. Yeah Yeah Yeah Yeah Yeah Yeah Yeah Yeah. Yeah Yeah Yeah'' \\
\bottomrule
\end{tabularx}
\end{table}

\paragraph{Silence.}
We split each run into one-minute bins and call a bin \emph{silence} if it
contains fewer than three words. Prolonged-silence onset is the first of two
consecutive silence bins, so brief pauses are not counted.

\paragraph{Gibberish.}
We measure gibberish in two ways. For benchmark results, Gemini-2.5-Flash
(temperature 0) labels each one-minute bin with at least three words as
gibberish or not. This label uses only the output text, so it applies to
every model. For the mechanism analyses (\cref{app:diagnostics}), we also use the model's \emph{predictive entropy}
$H_t = -\sum_i p_t(i)\log p_t(i)$, where $p_t$ is its distribution over the
next text token at frame $t$. High entropy means the model is uncertain about
what to say next. Because it requires the model's logits, which not every
evaluated model exposes, we do not use it for benchmark results. A bin is labeled gibberish if its text is largely uninterpretable English,
such as word salad, broken token fragments, or severely malformed syntax.
Repetition, disfluency, generic or off-topic replies, and factual errors do
not count as gibberish on their own. Gibberish onset is the first bin labeled
gibberish. The exact classification prompt is given in \cref{app:prompts}.


\paragraph{Coherence and Context.}
Two further Gemini calls score each speaking bin from 1 to 5. Coherence rates
whether the speech is fluent and makes sense; Context rates whether it
responds appropriately to the partner, so this call also sees the partner's
recent speech. Silence bins receive a score of 1, so silence lowers the
average instead of dropping out of it. The two scoring prompts are given in
\cref{app:prompts}.

\paragraph{Aggregation.}
\Cref{tab:main_results,tab:moshi_ablation_results} report how many runs reach each failure and, among those runs, the
mean${}\pm{}$SD onset time. Time-to-event curves(\cref{fig:candor_failure_free,fig:moshi_failure_free_ablation}) show the fraction of runs
that have not yet failed, estimated with the Kaplan--Meier
method~\citep{kaplan1958nonparametric}. Some runs end early because the model
reaches its context limit or the input audio ends. Such a run is \emph{right-censored}: we cannot know whether it would have
failed later, so it counts as failure-free while it is running and is left
out of the calculation after it ends. As a result, a curve's final value can differ slightly from the
raw fraction of failure-free runs in the tables. \textsc{NR} (not reached)
means the event did not occur before the run ended.

\subsection{Human Validation of the Gemini Gibberish Judge}
\label{app:human_validation}
\label{sec:human_gibberish_validation}

We checked the Gemini gibberish label against human listeners. Ten
proficient-English listeners were recruited through Prolific, gave informed
consent, and were paid for their time; two failed an attention check and were
excluded, leaving eight. Each rated 40 one-minute clips from Moshi and
PersonaPlex, half labeled gibberish by Gemini and half not. Listeners did not
see the model, condition, transcript, or Gemini label. A clip is gibberish for
humans if at least five of eight listeners say so.

Gemini agrees with the human majority on 36 of 40 clips (Cohen's
$\kappa=0.80$; 1 means perfect agreement), similar to agreement between
listeners (mean pairwise $\kappa=0.76$). In all four disagreements, humans
heard gibberish that Gemini missed; Gemini never called normal speech
gibberish. So the real amount of gibberish may be slightly higher than we
report (\cref{tab:human_gibberish_validation}).

\begin{table}[!ht]
\centering
\small
\setlength{\tabcolsep}{5pt}
\caption{\textbf{Human validation of the Gemini gibberish label.}
Eight listeners rated the same 40 clips; the human majority is the reference.}
\label{tab:human_gibberish_validation}
\begin{tabular}{ll}
\toprule
Measure & Result \\
\midrule
Agreement with human majority & 36/40 (90\%) \\
Cohen's $\kappa$ (Gemini vs.\ humans) & 0.80 (95\% CI 0.61--0.99) \\
Mean pairwise $\kappa$ (human vs.\ human) & 0.76 (28 pairs) \\
Precision (Gemini gibberish is correct) & 20/20 (1.00) \\
Recall (human gibberish found by Gemini) & 20/24 (0.83) \\
\bottomrule
\end{tabular}
\end{table}

Two caveats apply. First, humans listened to audio, while Gemini read the
transcript. Second, we picked the clips to be exactly half gibberish, but real
benchmark data does not have this even split, so the 90\% agreement may not
hold exactly on the full benchmark. Overall, the study shows that the Gemini
label is a reasonable proxy for human judgment.

\subsection{Checking the Collapse Metrics on Audio}
\label{app:audio_validation}
Our collapse metrics use the model's text output, because every model
produces text and this makes the metrics easy to reproduce. Here we check, on saved audio, that silence and repetition can also be heard in the speech
(\cref{tab:audio_metric_validation}). Gibberish is already checked
by human listeners in \cref{app:human_validation}.

\begin{table}[!ht]
\centering
\footnotesize
\setlength{\tabcolsep}{4pt}
\caption{\textbf{Checking the collapse metrics on audio.} For each failure,
we check whether the text-based metric agrees with the audio.}
\label{tab:audio_metric_validation}
\begin{tabularx}{0.9\linewidth}{@{}lXX@{}}
\toprule
Failure & How we check it on audio & What we find \\
\midrule
Silence
  & Measure how loud each minute is
  & Loudness and word count agree on which minutes are silent
    (AUROC 0.983) \\
Repetition
  & Transcribe the audio with Whisper and recompute repetition
  & Runs repetitive in text are also repetitive in audio; IAP+DRP also reduces
    repetition in the audio \\
\bottomrule
\end{tabularx}
\end{table}

\paragraph{Silence.}
We split each one-minute bin into 20-ms frames and count a frame as
active if it is louder than $-45$ dBFS, a level close to silence. A bin
counts as silent if its share of active frames is below a cutoff. We use
half of the runs to choose this cutoff and the other half to test it. We
then check whether these silent bins match the bins that our word-count
definition calls silent. The two definitions agree closely. They label
almost the same minutes as silent (AUROC 0.983). Their silence onsets
differ by only 0.67 minutes on average. Both definitions also find the
same runs with prolonged silence.

\paragraph{Repetition.}
The model's text output may not exactly match what it says aloud, so we
also measure repetition from the audio. We transcribe all 160 replay runs of
Moshi and PersonaPlex (Baseline and IAP+DRP) with two ASR models, Whisper
\texttt{small.en} and \texttt{large-v3}~\citep{radford2023whisper}, and apply
the same repetition criterion to each transcript
(\cref{tab:audio_repetition}). When the text shows repetition, the audio
almost always does too (precision 1.00 with \texttt{small.en} and 0.98 with
\texttt{large-v3}). The text-based metric also misses some audible repetition: about 30\% of
the runs flagged from audio are not flagged from text (recall 0.72 and
0.68), so it tends to undercount repetition.
Even when repetition is measured from the audio, IAP+DRP still greatly
reduces it compared with the baselines.

\begin{table}[!ht]
\centering
\small
\caption{\textbf{Repetition detected from text and from audio.} Number of
replay runs (out of 40) that reach repetition, using the model's text output
or ASR transcripts of its audio.}
\label{tab:audio_repetition}
\begin{tabular}{llccc}
\toprule
Model & Method & Text output & Whisper \texttt{small.en} & Whisper \texttt{large-v3} \\
\midrule
Moshi       & Baseline & 26 & 40 & 40 \\
Moshi       & IAP+DRP  & 0  & 2  & 8  \\
PersonaPlex & Baseline & 36 & 40 & 40 \\
PersonaPlex & IAP+DRP  & 1  & 5  & 3  \\
\bottomrule
\end{tabular}
\end{table}


\subsection{Why Predictive Entropy Instead of Perplexity}
\label{app:metric_sensitivity}
\label{sec:ppl_coherence_audit}
A common way to score generated speech is the perplexity (PPL) of its
transcript under an external language model~\citep{lakhotia2021generative},
and full-duplex models, including Moshi, have been evaluated this
way~\citep{defossez2024moshi,veluri2024beyond}. However, PPL measures how
predictable text is, not whether it makes sense. A repetitive loop is highly
predictable and gets low PPL~\citep{holtzman2019curious,park2024long}, while
unusual but readable speech can get high PPL.

\begin{figure}[!ht]
\centering
\includegraphics[width=\linewidth]{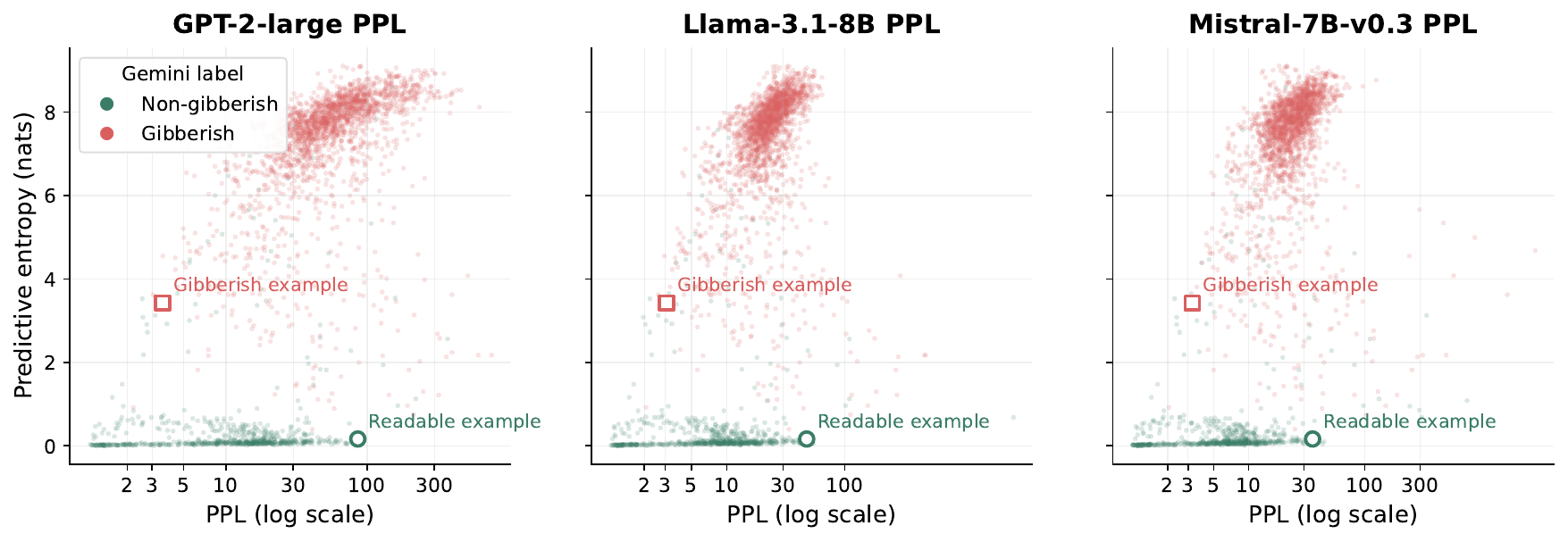}
\caption{\textbf{PPL does not separate gibberish, while predictive entropy
does.} Each point is a speaking bin, colored by the Gemini label (red:
gibberish, green: non-gibberish). Along the x-axis (PPL, log scale), the two
colors overlap; along the y-axis (predictive entropy), they separate clearly.
Marked points are examples from \cref{tab:ppl_gibberish_examples}.}
\label{fig:gibberish_metric_comparison}
\end{figure}

\Cref{tab:ppl_gibberish_examples} shows examples where readable speech gets
\emph{higher} PPL than gibberish under three external LMs:
GPT-2-large~\citep{radford2019language},
Llama-3.1-8B~\citep{grattafiori2024llama}, and
Mistral-7B-v0.3~\citep{jiang2023mistral7b}.
\Cref{fig:gibberish_metric_comparison} shows this over 2,750 speaking bins.
Each point is one bin, placed by its PPL (x-axis) and its predictive entropy
(y-axis). Along the x-axis, gibberish (red) and non-gibberish (green) bins
cover overlapping PPL ranges, so no single PPL cutoff separates them. Some
gibberish bins even get lower PPL than typical readable speech, as the two
marked examples from \cref{tab:ppl_gibberish_examples} show. Along the
y-axis, the two groups barely overlap: non-gibberish bins stay near zero
entropy, while most gibberish bins lie far above. The same pattern holds for
all three external LMs. We therefore use the Gemini label for benchmark
results and predictive entropy for mechanism analyses. This does not mean PPL
is useless in general, only that it does not detect gibberish here.

\begin{table*}[!htbp]
\centering
\scriptsize
\setlength{\tabcolsep}{2.7pt}
\renewcommand{\arraystretch}{1.08}
\caption{\textbf{Readable speech can get higher PPL than gibberish.}
Eight example bins from Moshi and PersonaPlex. Under all three external LMs,
the readable examples get higher PPL than the gibberish ones, because
gibberish here often consists of short, repeated fragments that are easy to
predict. Predictive entropy $H$ and the Gemini label $G$ (1: gibberish)
separate them correctly. Excerpts are shortened; \cref{fig:gibberish_metric_comparison}
shows all bins.}
\label{tab:ppl_gibberish_examples}
\begin{tabularx}{\textwidth}{@{}llccccc>{\raggedright\arraybackslash}X@{}}
\toprule
Type & Model & GPT-2 PPL & Llama PPL & Mistral PPL & $H$ & $G$ & Assistant-text excerpt \\
\midrule
\multirow{4}{*}{Readable}
& Moshi & 68.9 & 45.0 & 42.4 & 0.07 & 0 &
``Hello, what's up? I was just talking to my mom earlier, and I got this really good idea. How about yours?'' \\
& Moshi & 36.8 & 30.5 & 19.1 & 0.11 & 0 &
``I caught up on some reading. That sounds like a good book. I should check it out.'' \\
& PersonaPlex & 86.0 & 47.6 & 35.5 & 0.16 & 0 &
``Hello. My name's Angela. I'm in Ohio. Are you near Orlando? What is your opinion?'' \\
& PersonaPlex & 53.5 & 36.0 & 22.3 & 0.12 & 0 &
``My week's been okay. It's been busy. How about yours? I'm finding out as a teacher, it's true.'' \\
\midrule
\multirow{4}{*}{Gibberish}
& Moshi & 4.5 & 3.3 & 3.6 & 3.81 & 1 &
``then a a a a a a then a a a~\ldots{} worldend a a a then a grand~\ldots'' \\
& Moshi & 5.2 & 4.2 & 4.7 & 6.54 & 1 &
``a a-- at the Under a a a again~\ldots{} a behind a~\ldots{} a a a a a~\ldots'' \\
& PersonaPlex & 3.5 & 3.1 & 3.3 & 3.42 & 1 &
``of many of of of of a~\ldots{} of four of of all~\ldots{} c c of~\ldots'' \\
& PersonaPlex & 5.0 & 3.7 & 3.9 & 4.39 & 1 &
``of these large of these probably of of~\ldots{} Johnson~\ldots{} of of of next~\ldots'' \\
\bottomrule
\end{tabularx}
\end{table*}

\subsection{Sensitivity to Metric Thresholds}
\label{app:threshold_sensitivity}

\paragraph{Repetition.}
Our repetition criterion involves three choices: the Seq-Rep-4 threshold
(0.30), the window size (32 words), and the persistence rule (two consecutive
windows). To check that our conclusions do not hinge on these choices, we
recompute repetition on the 20 longest replays while varying each one:
thresholds of 0.20, 0.30, and 0.40; windows of 16, 32, and 64 words; and a
single-window rule that flags a run as soon as one window crosses the
threshold. \Cref{tab:repetition_sensitivity} reports how many of the 20 runs
are flagged in each setting. Looser settings flag more runs, as expected. In
every setting, however, the baselines are flagged in at least 9 of 20 runs,
whereas IAP+DRP is flagged in no run under the two-window rule and in at most
3 runs under the single-window rule. The gap between the baselines and
IAP+DRP therefore does not depend on these choices.
\begin{table}[!htbp]
\centering
\footnotesize
\setlength{\tabcolsep}{4pt}
\caption{\textbf{Repetition sensitivity on the 20 longest replays.} Each cell
gives the number of runs (out of 20) flagged as repetitive at thresholds
0.20/0.30/0.40. Windows advance by half their length and cover at most
30 seconds. Our main setting is two consecutive 32-word windows at 0.30 (the
middle number in the 32-word column of the upper block). The subset was fixed
before examining outcomes.}
\label{tab:repetition_sensitivity}
\begin{tabular}{@{}lllccc@{}}
\toprule
Rule & Model & Method & 16 words & 32 words & 64 words \\
\midrule
two consecutive & Moshi & Baseline & $19/14/9$ & $17/14/11$ & $18/12/10$ \\
two consecutive & Moshi & IAP+DRP & $0/0/0$ & $0/0/0$ & $0/0/0$ \\
two consecutive & PersonaPlex & Baseline & $20/19/12$ & $20/18/14$ & $19/17/13$ \\
two consecutive & PersonaPlex & IAP+DRP & $0/0/0$ & $0/0/0$ & $0/0/0$ \\
\midrule
single window & Moshi & Baseline & $20/19/15$ & $20/18/13$ & $18/14/10$ \\
single window & Moshi & IAP+DRP & $0/0/0$ & $0/0/0$ & $0/0/0$ \\
single window & PersonaPlex & Baseline & $20/20/19$ & $20/20/17$ & $20/19/17$ \\
single window & PersonaPlex & IAP+DRP & $3/2/0$ & $1/0/0$ & $0/0/0$ \\
\bottomrule
\end{tabular}
\end{table}

\paragraph{Gibberish.}
Unlike repetition and silence, gibberish onset requires only a single bin
labeled gibberish. To check whether this choice matters, we also apply a
stricter rule that requires three consecutive speaking bins labeled
gibberish. For the two sliding-window models, the comparison between the
baseline and IAP+DRP is unchanged. In replay, both baselines still reach
gibberish in 40 of 40 runs and IAP+DRP in none. In interaction, the baselines
drop only slightly (Moshi from 40 to 38 runs, PersonaPlex from 39 to 35),
while IAP+DRP goes from 2 to 1 run for Moshi and stays at 0 for PersonaPlex.
The stricter rule matters for the growing-context models. BayLing-Duplex
often falls silent right after its first gibberish bin, so its count drops
from 36 to 3 runs in replay and from 40 to 11 in interaction; its collapse
then shows up as silence instead. Lychee-FD never produces three gibberish
bins in a row, so its count drops to 0 in both tracks. A three-bin rule would
therefore hide most gibberish in these models, and we use the single-bin
rule.

\subsection{Gemini Prompts}
\label{app:prompts}
The following are the prompt templates used for the Gemini-based measures and
the interactive partner. Braced fields denote text inserted at run time. The
three evaluation prompts use Gemini~2.5~Flash at temperature 0. The partner
uses the same model at temperature 0.9, receives the last 16 dialogue turns,
and generates at most 90 tokens per turn. Judge inputs are JSON-encoded; the
assistant text is truncated to 1,500 characters and, for Context, the user
text to 1,200 characters.

\paragraph{Binary gibberish classification.}
\begin{Verbatim}[breaklines=true,breakanywhere=true,fontsize=\scriptsize,frame=single]
Classify whether the assistant text is linguistic gibberish.

GIBBERISH = 1:
The output is largely uninterpretable as English because of word salad,
broken token fragments, severely malformed syntax, or nonsensical word
composition.

GIBBERISH = 0:
The output remains understandable English, even if it is repetitive,
disfluent, generic, factually incorrect, or unrelated to the user.

Judge only the assistant text. Do not evaluate relevance. Do not classify
understandable repetition alone as gibberish.

ASSISTANT TEXT: {assistant_text_json}
Return ONLY JSON: {"gibberish": 0 or 1}
\end{Verbatim}

\paragraph{Coherence score.}
\begin{Verbatim}[breaklines=true,breakanywhere=true,fontsize=\scriptsize,frame=single]
Rate ONLY the intrinsic fluency of this one-minute assistant speech segment
on an integer scale from 1 to 5. Read only the assistant text.

5 = fluent, grammatical, and sensible English, even if unrelated to the user.
3 = clunky or somewhat repetitive, but still readable and interpretable.
1 = gibberish, word salad, broken fragments, or speech stuck in a looping
pattern.

ASSISTANT: {assistant_text_json}
Return ONLY JSON: {"coherence": n}
\end{Verbatim}

\paragraph{Context score.}
\begin{Verbatim}[breaklines=true,breakanywhere=true,fontsize=\scriptsize,frame=single]
Rate ONLY how appropriately this one-minute assistant speech segment responds
to the user's speech from the matched minute, on an integer scale from 1 to 5.

5 = on-topic and relevant; short backchannels can be highly appropriate.
3 = vague or generic, but plausibly responsive.
1 = irrelevant, canned, or ignores the user. Uninterpretable gibberish cannot
be relevant and must score 1.

USER: {user_text_json}
ASSISTANT: {assistant_text_json}
Return ONLY JSON: {"context": n}
\end{Verbatim}

\paragraph{FD-LHB-LLM partner.}
The system instruction is:
\begin{Verbatim}[breaklines=true,breakanywhere=true,fontsize=\scriptsize,frame=single]
You are a warm, friendly person on a casual phone call with a friend named
Alex. Talk naturally about everyday life -- your day, hobbies, food, weekend
plans, whatever comes up. React genuinely to what Alex says, ask follow-up
questions, share your own little stories. Keep each turn SHORT (1-2
sentences), like real speech. Do NOT quiz them or recite facts -- just a nice
flowing chat. If Alex says something odd, unclear, or repetitive, react like a
real person would (e.g. 'wait, what?', 'you okay?', gently change topic). This
particular call naturally revolves around {topic}.
\end{Verbatim}
For the first turn, before any dialogue history is available, the user message
is:
\begin{Verbatim}[breaklines=true,breakanywhere=true,fontsize=\scriptsize,frame=single]
(start the call with a warm, natural greeting, then soon bring up {topic})
\end{Verbatim}

\subsection{Reproducibility}
\label{app:reproduction}
FD-LHB involves no training: it runs existing models on fixed inputs and
scores their outputs. Upon publication, we will release the code to run each
model with IAP and DRP, the FD-LHB-LLM partner, and the metric computation,
including the Gemini judge prompts. We will also release a manifest listing
the CANDOR conversations and crop boundaries used in each experiment. CANDOR
audio is available only through its research-access process, so we release
the manifest rather than the audio. To reproduce the results, one rebuilds the
inputs from the manifest, runs each model with the settings in
\cref{tab:repro_settings}, and scores the outputs as described in
\cref{app:metrics}.
\FloatBarrier

\section{Diagnostic Experiments: Details and Additional Analyses}
\label{app:diagnostics}
This section gives the setup of the diagnostic experiments in the main text
and additional analyses of the two mechanisms: initialization-anchor
eviction (\cref{app:anchor_analysis}) and self-reinforcing repetition
(\cref{app:repetition_analysis}). The anchor analyses measure gibberish with
predictive entropy (\cref{app:metrics}) and a threshold of $H=2$ nats,
rather than with the Gemini label.

\subsection{Initialization-Anchor Eviction}
\label{app:anchor_analysis}
\label{sec:state_drift}
\label{sec:pinned_identity}
\begin{figure}[t]
\centering
\includegraphics[width=0.8\linewidth]{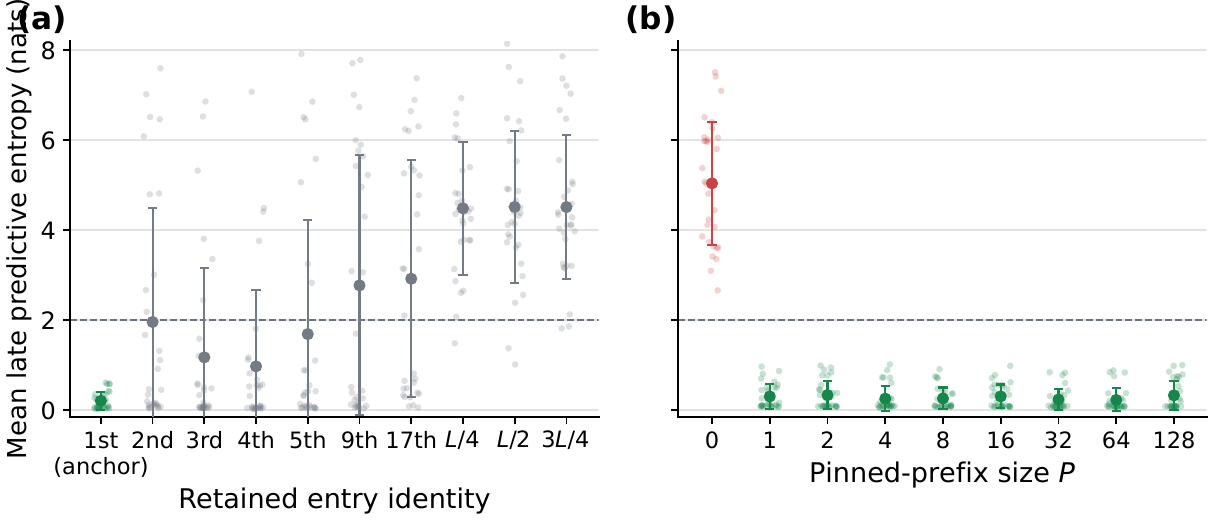}
\caption{\textbf{Late-conversation entropy when keeping different cache
entries.} (a) Keeping one early entry. (b) Keeping the first $P$ entries.
Each dot is one run; the dashed line is $H=2$.}
\label{fig:anchor_identity_entropy_pooled}
\end{figure}
\paragraph{Evidence for the training premise.}
The main text assumes that every training sequence starts from an empty
cache. We confirmed this in the official Moshi fine-tuning
code~\citep{moshifinetune}, where each training sample is a segment of at
most \texttt{duration\_sec} seconds processed from an empty cache. Its README
also warns that lowering \texttt{duration\_sec} may cause the model to become
silent more quickly at inference, which is consistent with a mismatch between
training and inference lengths. This code covers fine-tuning rather than
pretraining, and PersonaPlex is trained separately, so we test the role of
anchor eviction directly with the interventions below.

\paragraph{Only the first entry matters.}
In the main text, we test which cache entries must be kept to prevent
gibberish (\cref{fig:anchor_identity_count}). Here we describe these tests in
more detail and show the entropy of every run. Both tests use 15 CANDOR
conversations for each of Moshi and PersonaPlex (30 runs per condition, fixed
decoding seed). The \emph{identity} test keeps one early entry and varies only
which one it is, with its cache slot and position index fixed; the
\emph{count} test keeps the first $P$ entries,
$P\in\{0,1,2,4,\dots,128\}$. \Cref{fig:anchor_identity_entropy_pooled} shows each run's entropy averaged
over the late part of the conversation (minutes 13--20 in (a) and 10--30 in
(b)), together with the mean and SD of each condition. In the identity test (a), only the first entry
keeps every run below $H=2$; every other choice leaves some runs far above it,
so the result does not depend on the exact threshold. In the count test (b),
entropy is high without any retained entry ($P=0$) and low for every
$P\geq1$. Varying the decoding seed (42--46) on one PersonaPlex replay gives
the same result.

\begin{figure*}[!htbp]
\centering
\includegraphics[width=0.92\textwidth]{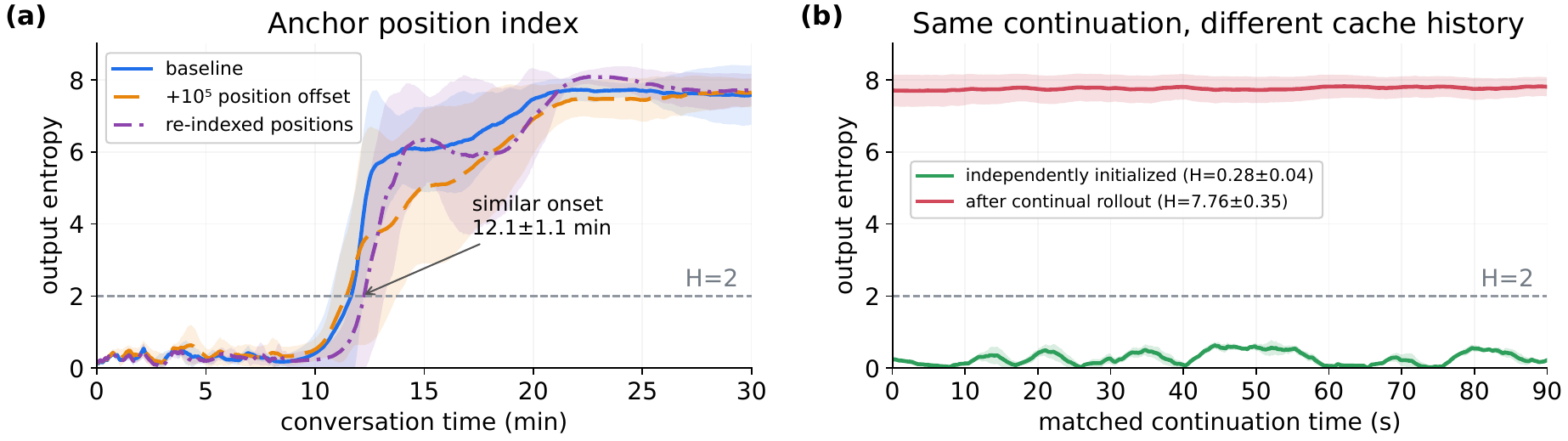}
\caption{\textbf{Position indices do not matter; the cache history does.}
(a) Output entropy over a conversation with the original, shifted, or
re-indexed position indices. (b) Output entropy for the same 90-second audio,
played after a fresh start or after a long conversation. Mean ${}\pm{}$ SD
over five seeds, pooled across Moshi and PersonaPlex.}
\label{fig:position_cache_history}
\end{figure*}

\paragraph{Large position indices are not the cause.}
In long conversations, position indices grow far beyond those seen in
training. Because RoPE depends only on relative distance, this should not
matter, and \cref{fig:position_cache_history}a confirms it. Shifting all
positions by $10^5$, or re-indexing the visible cache at every step so that
its positions always run from $0$ to $L-1$ and never exceed the training
range, leaves the onset of high entropy unchanged ($12.1\pm1.1$ minutes).
\begin{wrapfigure}[14]{r}{0.60\linewidth}
\vspace{-0.4em}
\centering
\includegraphics[width=0.98\linewidth]{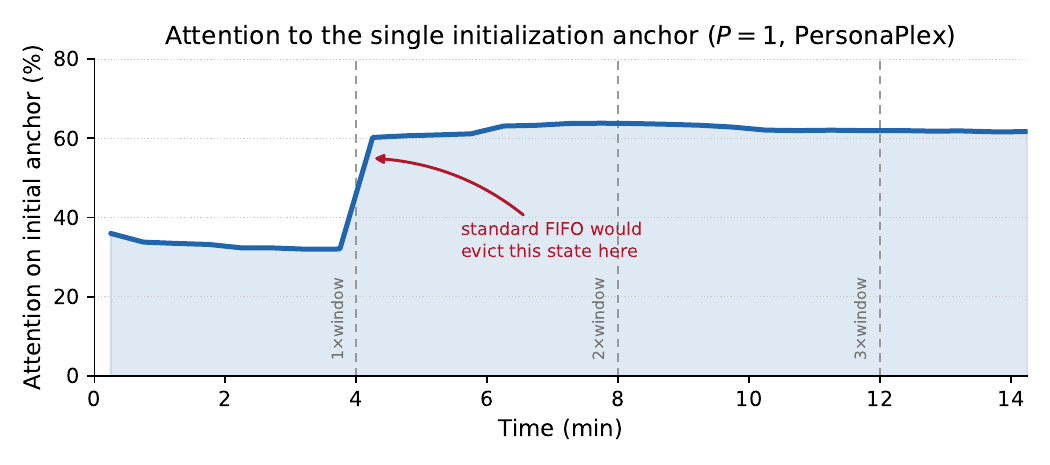}
\vspace{-0.5cm}
\caption{\textbf{The pinned anchor acts as an attention sink.} Share of
attention on the pinned anchor in one PersonaPlex replay with IAP ($P=1$),
in 30-second averages. Dashed lines mark each full turnover of the sliding window.}
\label{fig:attn_sink}
\end{wrapfigure}
\paragraph{The cache history is what matters.}
If position indices are not the cause, what differs between early and late
in a conversation is the history stored in the cache. To isolate it, we play
the same 90-second audio to the model twice: once right after a fresh start,
with an empty cache, and once after the model has already run a long
conversation. Everything else is identical, including the audio, the
position indices, and the decoding settings. After a fresh start, entropy
stays low ($H=0.28\pm0.04$); after the long conversation, it is very high
($H=7.76\pm0.35$; \cref{fig:position_cache_history}b). The history in the
cache therefore changes the model's behavior.

\paragraph{The pinned anchor acts as an attention sink.}
IAP resembles StreamingLLM~\citep{xiao2024efficient}, which also keeps the
first few tokens of a sequence in the cache. StreamingLLM showed that language
models place a large share of their attention on these first tokens, which it
calls \emph{attention sinks}, and that the model fails once they are removed.
To check whether the pinned anchor plays the same role, we measure how much
attention it receives over time. We run IAP ($P=1$) on one 14-minute
PersonaPlex replay and record the share of attention that goes to the pinned
anchor, averaged over all layers and heads (\cref{fig:attn_sink}). The anchor
already receives about 32\% of the attention during the first window. After
the first window turnover at 4 minutes, its share rises above 60\% and stays there. The pinned anchor therefore
acts as an attention sink, and the model relies on it even more once the rest
of the early context has left the window.

\subsection{Self-Reinforcing Repetition}
\label{app:repetition_analysis}
\label{sec:self_reinforcement}

\paragraph{Word-insertion experiment.}
This section gives the setup of the word-insertion experiment in
\cref{sec:self_reinforcing_repetition} (\cref{fig:self_reinforcing_repetition}c).
For each model, we take 20 CANDOR contexts that later reach repetition and
branch them at least 30 seconds before repetition onset. Each branch
receives $d\in\{0,1,2,4,8,16,32\}$ forced occurrences of \emph{purple}, a
word absent from the preceding two minutes, and then runs freely for 20
seconds with three sampling seeds. We then measure how often \emph{purple} appears among the model's words
in these 20 seconds. Forced occurrences are not counted.
Without insertion, neither model says \emph{purple}. In PersonaPlex, its
share rises sharply up to $d=4$ and keeps growing, reaching 28.4\% at
$d=32$. In Moshi, it stays below 10\% at every dose.

\section{Implementation Details of IAP and DRP}
\label{app:implementation}
This section provides the implementation details and final settings of IAP
and DRP. Evaluation results and ablations are reported in
\cref{app:ablations}.

\subsection{Initialization-Anchor Pinning}
\label{app:iap}
IAP pins the first KV entry ($P=1$) in the cache. The other $L-1$ slots
follow the normal FIFO update, so the cache size stays at $L$ and no
training is needed.

The pinned entry needs one extra step. RoPE~\citep{su2023roformerenhancedtransformerrotary}
attention depends on the distance between query and key positions. If the
pinned entry stayed at its original position, its distance to new queries
would keep growing and soon exceed the
distances seen in training. IAP
therefore increases the pinned entry's position by one each time the window
slides, so that it always sits right before the oldest remaining entry. All
other entries keep their positions, and each new entry receives the next
position. To do this, we store the pinned key without RoPE applied and re-apply RoPE
at its new position whenever it moves. Only this one key is updated at each
step, instead of re-indexing the whole cache as in
StreamingLLM~\citep{xiao2024efficient}.

\subsection{Duplex Repetition Penalty}
\label{app:drp}

DRP modifies the text logits in two parts. The first penalizes recently used
lexical tokens while keeping the total lexical probability unchanged, so
that the penalty does not make the model less likely to speak. The second
applies a small \texttt{PAD} correction after unusually long silent periods.
DRP is applied before temperature scaling and top-$k$ filtering. When
predictive entropy is used for analysis, it is computed from the original
logits before applying DRP.

\paragraph{Penalizing recent lexical tokens.}
Let $z_{t,i}$ be the text logit of token $i$ at frame $t$, let $\mathcal{V}$
be the set of lexical text tokens (all text tokens except \texttt{PAD} and
\texttt{EPAD}), and let $\mathcal{H}_t\subseteq\mathcal{V}$ be the set of
tokens appearing among the model's last $N$ lexical tokens. We call $N$ the history length. We first apply
the repetition penalty of~\citet{keskar2019ctrl}:
\begin{equation}
R_\rho(z_{t,i}) =
\begin{cases}
z_{t,i}/\rho, & i\in\mathcal{H}_t,\ z_{t,i}>0,\\
\rho z_{t,i}, & i\in\mathcal{H}_t,\ z_{t,i}\leq0,\\
z_{t,i}, & \text{otherwise},
\end{cases}
\label{eq:drp_repetition_penalty}
\end{equation}
where $\rho>1$ controls the penalty strength. Each token in $\mathcal{H}_t$
receives the same penalty regardless of how often it appeared.

This penalty also lowers the total probability of lexical tokens, which
would make \texttt{PAD} and \texttt{EPAD} relatively more likely. To prevent
this, we then add the same offset $c_t$ to every lexical-token logit:
\begin{equation}
\begin{aligned}
\tilde z_{t,i} &= R_\rho(z_{t,i}) + c_t, \qquad i\in\mathcal{V},\\[4pt]
c_t &= T_t \log
\frac{\sum_{j\in\mathcal{V}} \exp\!\big(z_{t,j}/T_t\big)}
     {\sum_{j\in\mathcal{V}} \exp\!\big(R_\rho(z_{t,j})/T_t\big)},
\end{aligned}
\label{eq:drp_lexical_mass}
\end{equation}
where $T_t$ is the text temperature. With this offset, the total lexical
probability after temperature scaling is the same as before the penalty.
Probability removed from recently used tokens therefore goes to other
lexical tokens rather than to \texttt{PAD} or \texttt{EPAD}.

\paragraph{Prolonged-\texttt{PAD} safeguard.}
The repetition penalty can occasionally make it harder for the model to
resume speaking after a long silence. We therefore lower the \texttt{PAD}
logit when the model has stayed silent for a long time and VAD indicates
that the user is quiet.

Let $r_t$ be the current number of consecutive \texttt{PAD} frames, $r_0$ the
length at which the correction begins, and $\tau_{\mathrm{ramp}}$ the number
of frames over which it grows to its maximum $b_{\max}$. Let $q_t=1$ when VAD
detects that the user is quiet and $q_t=0$ otherwise. At frame $t$, we
subtract
\begin{equation}
b_t
=
q_t b_{\max}
\operatorname{clip}\!\left(
\frac{r_t-r_0+1}{\tau_{\mathrm{ramp}}},
0,1
\right)
\label{eq:drp_pad_control}
\end{equation}
from the \texttt{PAD} logit, where $\operatorname{clip}(x,0,1)=\min(\max(x,0),1)$.
The correction is zero until the silence reaches $r_0$ frames, then grows
linearly over $\tau_{\mathrm{ramp}}$ frames up to $b_{\max}$. It is off while
the user is speaking and resets when the model outputs \texttt{EPAD} or a
lexical token.

\subsection{Final Settings}
\label{app:final_configuration}

Our final configuration uses a history length of $N = 512$ lexical tokens.
For the prolonged-\texttt{PAD} safeguard, the correction begins when the
\texttt{PAD} run reaches $r_0=375$ frames (30\,s), increases to its maximum
over $\tau_{\mathrm{ramp}}=125$ frames (10\,s), and is capped at
$b_{\max}=3.0$. The repetition-penalty strength is $\rho=1.5$ for Moshi
and $\rho=1.3$ for PersonaPlex. IAP retains a single initialization anchor
($P=1$).

Model-specific generation settings are summarized
in \cref{tab:repro_settings}.

\begin{table}[!htbp]
\centering
\footnotesize
\setlength{\tabcolsep}{3pt}
\renewcommand{\arraystretch}{1.08}
\caption{\textbf{Generation settings for each model.}
$T_a/T_t$ denote audio/text temperatures and $k_a/k_t$ the corresponding
top-$k$ values. The last column reports the context limit and evaluation
horizon. Baseline and Ours use identical settings within each track.}
\label{tab:repro_settings}
\resizebox{\linewidth}{!}{%
\begin{tabular}{@{}lllll@{}}
\toprule
System and track & $T_a/T_t$ & Sampling restriction & Seed & Context / evaluation horizon \\
\midrule
PersonaPlex, both tracks & $.8/.65$ & $k_a/k_t=250/25$ & 42 & 3,000 frames / full replay; 20 min interaction \\
PersonaPlex, diagnostics & $.8/.7$ & $k_a/k_t=250/25$ & 42--46 & varied window / 20 or 30 min \\
Moshi, replay (Baseline and Ours) & $.8/.7$ & $k_a/k_t=250/25$ & 42 & 3,000 frames / full input \\
Moshi, interaction (Baseline and Ours) & $.8/.9$ & $k_a/k_t=250/25$ & 42 & 3,000 frames / 20 min \\
Moshi, diagnostics & $.8/.7$ & $k_a/k_t=250/25$ & 42--46 & varied window / 20 or 30 min \\
Raon-SpeechChat, replay & --/.6 & top-$p=.8$, top-$k=20$, BC${}=2$ & 42 & 8,192-position cache cap \\
Raon-SpeechChat, interaction & --/.9 & top-$p=.95$, top-$k=66$, BC${}=0$ & 42 & 8,192-position cache cap / 20 min \\
BayLing-Duplex, replay & --/.6 & top-$p=.8$; ratio $10{:}5{:}10$ & 42 & 32,768-position native cap \\
BayLing-Duplex, interaction & --/.6 & top-$p=.8$; ratio $10{:}5{:}10$ & 42--81 & native cap / 20 min \\
Lychee-FD, both tracks & native API & native API & not exposed & 400-ms incremental / 10.92-min configured support \\
\bottomrule
\end{tabular}%
}
\end{table}

\section{Extended Results}
\label{app:extended_results}
\label{app:ablations}
This section extends the results in \cref{sec:results}.
\Cref{app:main_results} gives the complete benchmark results and
significance tests. \Cref{app:trajectories} reports response quality and
failure trajectories over long replays, and \cref{app:silence} examines how
long silences last. \Cref{app:components} measures the effect of IAP and DRP
separately, and \cref{app:drp_ablation} examines what each part of DRP
contributes. \cref{app:alternative_controls} replaces DRP's repetition penalty
with other repetition controls, \cref{app:hyperparameters} varies its main
hyperparameters, and \cref{app:human_tradeoff} checks with human listeners
whether DRP makes ordinary speech worse. Finally,
\cref{app:efficiency} reports the computational cost.

\subsection{Complete Main Results}
\label{app:main_results}
\label{sec:complete_numerical_results}

\begin{table}[t!]
\centering
\small
\setlength{\tabcolsep}{5pt}
\renewcommand{\arraystretch}{1.08}
\providecommand{\onsetcell}[2]{\makecell{#1\\[-1pt]{\scriptsize #2 reached}}}
\caption{\textbf{Complete FD-LHB results.} Every condition contains 40 runs.
$T_{\mathrm{rep}}$, $T_{\mathrm{psil}}$, and $T_{\mathrm{gib}}$ denote the
first of two consecutive repetitive 32-word windows, the first of two
consecutive silent one-minute bins, and the first gibberish-labeled
speaking bin, respectively. Onsets are mean $\pm$ sample SD over affected
runs. Configured or native context limits are right-censored; \NR{} means no
event was observed within the evaluation horizon.}
\label{tab:main_results}
\resizebox{0.88\linewidth}{!}{%
\begin{tabular}{@{}lllccc@{}}
\toprule
Track & Model & Method & $T_{\mathrm{rep}}$ & $T_{\mathrm{psil}}$ & $T_{\mathrm{gib}}$ \\
\midrule
\multirow{7}{*}{Replay}
& Raon-SpeechChat & Baseline & \onsetcell{\NR}{$0/40$} & \onsetcell{$3.2\pm1.1$}{$14/40$} & \onsetcell{$3.0$}{$1/40$} \\
& BayLing-Duplex & Baseline & \onsetcell{$2.2\pm1.6$}{$22/40$} & \onsetcell{$6.8\pm1.4$}{$40/40$} & \onsetcell{$4.2\pm1.2$}{$36/40$} \\
& Lychee-FD & Baseline & \onsetcell{$2.4\pm2.2$}{$15/40$} & \onsetcell{\NR}{$0/40$} & \onsetcell{$5.2\pm3.4$}{$12/40$} \\\cdashline{2-6}[0.4pt/1.2pt]
& PersonaPlex & Baseline & \onsetcell{$10.3\pm7.1$}{$36/40$} & \onsetcell{$8.2\pm1.3$}{$34/40$} & \onsetcell{$12.9\pm3.1$}{$40/40$} \\
& PersonaPlex + Ours & IAP+DRP & \onsetcell{$25.6$}{$1/40$} & \onsetcell{$31.6\pm15.1$}{$19/40$} & \onsetcell{\NR}{$0/40$} \\
& Moshi & Baseline & \onsetcell{$19.6\pm7.5$}{$26/40$} & \onsetcell{$9.0\pm0.8$}{$40/40$} & \onsetcell{$13.3\pm0.9$}{$40/40$} \\
& Moshi + Ours & IAP+DRP & \onsetcell{\NR}{$0/40$} & \onsetcell{$30.3\pm16.0$}{$3/40$} & \onsetcell{\NR}{$0/40$} \\
\midrule
\multirow{7}{*}{Interaction}
& Raon-SpeechChat & Baseline & \onsetcell{\NR}{$0/40$} & \onsetcell{\NR}{$0/40$} & \onsetcell{\NR}{$0/40$} \\
& BayLing-Duplex & Baseline & \onsetcell{$2.7$}{$1/40$} & \onsetcell{$8.2\pm2.7$}{$40/40$} & \onsetcell{$4.3\pm2.4$}{$40/40$} \\
& Lychee-FD & Baseline & \onsetcell{\NR}{$0/40$} & \onsetcell{\NR}{$0/40$} & \onsetcell{$3.2\pm3.3$}{$9/40$} \\\cdashline{2-6}[0.4pt/1.2pt]
& PersonaPlex & Baseline & \onsetcell{$7.1\pm4.6$}{$35/40$} & \onsetcell{$7.6\pm1.6$}{$35/40$} & \onsetcell{$11.8\pm2.0$}{$39/40$} \\
& PersonaPlex + Ours & IAP+DRP & \onsetcell{$16.9\pm3.1$}{$2/40$} & \onsetcell{$14.8\pm2.5$}{$4/40$} & \onsetcell{\NR}{$0/40$} \\
& Moshi & Baseline & \onsetcell{$10.0\pm6.1$}{$5/40$} & \onsetcell{$9.5\pm0.5$}{$28/40$} & \onsetcell{$12.2\pm0.8$}{$40/40$} \\
& Moshi + Ours & IAP+DRP & \onsetcell{\NR}{$0/40$} & \onsetcell{\NR}{$0/40$} & \onsetcell{$17.5\pm0.7$}{$2/40$} \\
\bottomrule
\end{tabular}%
}
\end{table}

\Cref{tab:main_results} gives the full numbers behind
\cref{fig:candor_failure_free}. Baseline and IAP+DRP receive the same 40 replays, so we compare their
results on each replay. We check whether IAP+DRP avoids each failure in more
replays than the baseline, using an exact McNemar test~\citep{mcnemar1947note}. This gives six
tests (two models and three failures). All six are significant, even with
the stricter Bonferroni threshold ($p<0.05/6$).

\subsection{Long-Horizon Trajectories and Response Quality}
\label{app:trajectories}
\label{sec:llm_judge_trajectories}

\begin{figure}[!htbp]
\centering
\includegraphics[width=\linewidth]{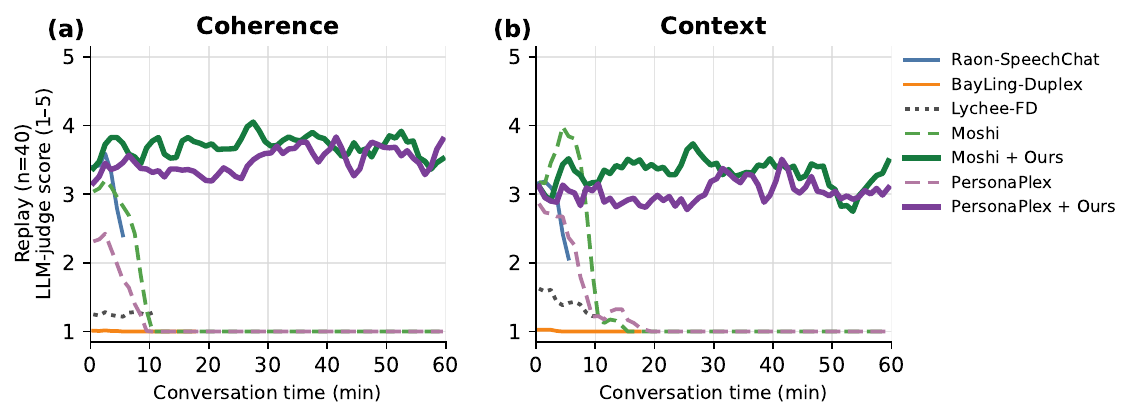}
\caption{\textbf{LLM-judge scores over time on FD-LHB-CANDOR.}
Mean (a) Coherence and (b) Context scores over 40 replays per system. Silence bins receive a score of 1. Curves for growing-context models end when the model
reaches its context limit.}
\label{fig:supp_llm_judge_quality}
\end{figure}

\Cref{fig:interaction_llm_judge} reports LLM-judge scores on FD-LHB-LLM; here
we report the same scores on the FD-LHB-CANDOR replays
(\cref{fig:supp_llm_judge_quality}). As in the main text, Gemini rates each
one-minute speaking bin from 1 to 5 for Coherence and Context, and silent
bins receive a score of 1 so that silence lowers the average instead of
dropping out of it. The replay results follow the same pattern as the
interaction track: both sliding-window baselines fall to the lowest score
within about 20 minutes, whereas IAP+DRP keeps both scores steady for the full
60 minutes. BayLing-Duplex stays near the lowest score throughout because it
often responds in a language other than English.

\begin{figure*}[!htbp]
\centering
\includegraphics[width=\textwidth]{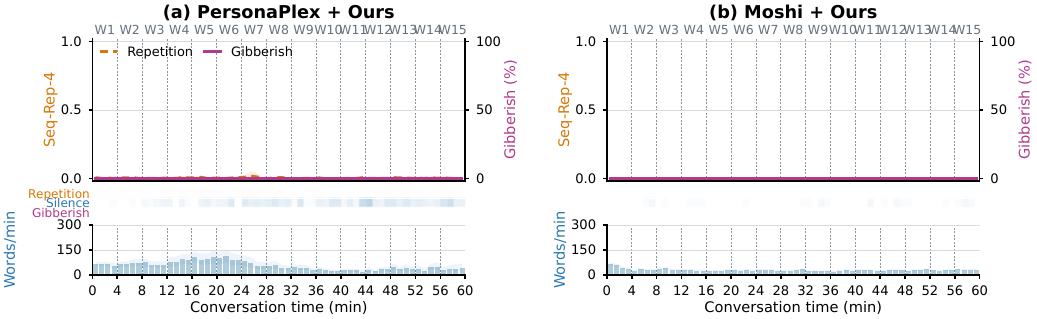}
\caption{\textbf{Failure trajectories with IAP+DRP over 40 CANDOR replays.}
Same format as \cref{fig:population_failure_patterns} for (a) PersonaPlex and
(b) Moshi with IAP+DRP, extended to 60 minutes. Top: repetition (Seq-Rep-4)
and the percentage of speaking bins labeled gibberish; middle: when each
failure appears; bottom: words per minute. W1--W15 mark consecutive
four-minute sliding windows.}
\label{fig:ours_full_trajectories}
\end{figure*}

\Cref{fig:ours_full_trajectories} repeats the view of
\cref{fig:population_failure_patterns} after applying IAP+DRP, extended to the
full 60-minute replay (W1--W15). Without IAP+DRP, both sliding-window models
reach gibberish about three windows in (\cref{fig:population_failure_patterns}a,b).
With IAP+DRP, repetition and gibberish stay near zero through all 15 windows,
and words per minute stay steady instead of rising sharply, as they do when the baselines turn to gibberish.

\subsection{Silence Duration and Resumption of Speech}
\label{app:silence}
\label{sec:silence_recovery}
The silence metric in the main text counts a run once it contains a
two-minute silence, even if the model later speaks again. Here we check how
long these silences last and what happens afterward. Over all 40 CANDOR
replays per condition, \cref{tab:silence_recovery} reports how many runs contain a silence of at least two minutes and of at least five minutes. It also reports how many runs resume speaking after the first two-minute silence, and how long this takes.
This checks only whether the model speaks again, not whether its speech is
coherent or relevant.
\begin{table}[!htbp]
\centering
\footnotesize
\setlength{\tabcolsep}{3.2pt}
\renewcommand{\arraystretch}{1.08}
\caption{\textbf{Silence duration and resumption of speech on all 40 CANDOR
replays.} A one-minute bin is silent if it contains fewer than three words;
the final partial minute is included. Any 2 min and Any 5 min count runs with
a silence of at least two or five consecutive minutes. Resumed counts runs
that produce at least one word after the first two-minute silence. Time to
resume is measured from the end of that silence to the first word; cells
report the median, IQR, and maximum in seconds.}
\label{tab:silence_recovery}
\resizebox{0.8\linewidth}{!}{%
\begin{tabular}{@{}llcccc@{}}
\toprule
Model & Method & Any 2 min & Any 5 min & Resumed &
Median time to resume (s) \\
\midrule
Moshi & Baseline & $40/40$ & $7/40$ & $40/40$ & \makecell{95.3\\[-1pt]{\scriptsize IQR 57.2--134.4; max 202.6}} \\
Moshi & IAP & $13/40$ & $0/40$ & $13/13$ & \makecell{7.5\\[-1pt]{\scriptsize IQR 2.6--24.8; max 70.2}} \\
Moshi & DRP & $37/40$ & $7/40$ & $37/37$ & \makecell{95.2\\[-1pt]{\scriptsize IQR 68.4--134.0; max 203.9}} \\
Moshi & IAP+DRP & $3/40$ & $0/40$ & $3/3$ & \makecell{10.2\\[-1pt]{\scriptsize IQR 6.7--16.1; max 22.0}} \\
PersonaPlex & Baseline & $34/40$ & $7/40$ & $34/34$ & \makecell{53.6\\[-1pt]{\scriptsize IQR 37.0--118.8; max 172.0}} \\
PersonaPlex & IAP+DRP & $19/40$ & $1/40$ & $18/19$ & \makecell{26.6\\[-1pt]{\scriptsize IQR 12.4--40.2; max 47.3}} \\
\bottomrule
\end{tabular}%
}
\end{table}

Two-minute silences occur in every Moshi baseline run and in 34 of 40
PersonaPlex baseline runs; IAP+DRP reduces them to 3 and 19 of 40,
respectively. In almost every condition, including the baselines, the model
eventually speaks again. The difference lies in how quickly the model resumes
and what it produces afterward. Without IAP, resumption is slow (median
95.3~s for Moshi and 53.6~s for PersonaPlex), and five-minute silences occur
in 7 of 40 runs. Moreover, although every baseline run resumes speaking, all
of them also reach gibberish, typically soon after resuming (silence starts at
about 9 minutes and gibberish at about 13 minutes; \cref{tab:main_results}).
This suggests that the silence is a transition into gibberish rather than a
pause. With IAP, resumption is much faster (median 7.5~s with IAP and 10.2~s
with IAP+DRP in Moshi, and 26.6~s with IAP+DRP in PersonaPlex), five-minute
silences almost disappear, and no run produces gibberish. In PersonaPlex, IAP+DRP still leaves a two-minute silence in 19 of 40 runs, but 18 of them resume speaking (median 26.6~s), so most of these silences are temporary pauses.

\subsection{Component Ablations}
\label{app:components}
\begin{figure}[!htbp]
\centering
\includegraphics[width=\linewidth]{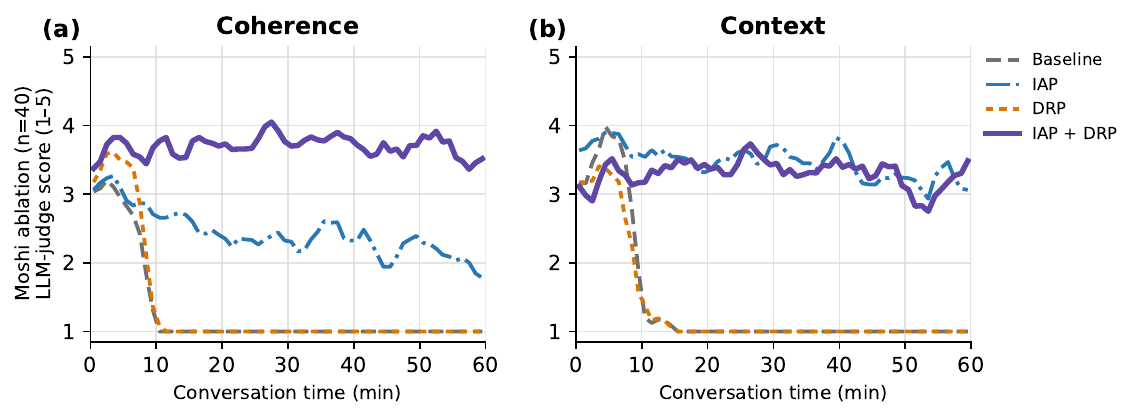}
\caption{\textbf{Response quality in the Moshi component ablation.}
Mean Coherence and Context scores over 40 matched CANDOR replays. Silence bins receive a score of 1.}
\label{fig:moshi_ablation_quality}
\end{figure}

\Cref{tab:moshi_ablation_results} compares Baseline, IAP, DRP, and IAP+DRP on
Moshi over 40 matched CANDOR replays. IAP alone removes gibberish in all 40
runs and also reduces silence (from 40 to 13 runs) and repetition (from 26 to
4 runs). DRP alone removes repetition in all runs but does not prevent
gibberish or silence (40 and 37 runs). Combining the two gives the best result
on all three failures, with silence in only 3 runs. In
\cref{fig:moshi_ablation_quality}, only IAP+DRP also keeps Coherence high over
the full 60 minutes. The two methods therefore play complementary roles,
although these results alone do not show that each failure has a single
cause.
\begin{table}[!h]
\centering
\small
\setlength{\tabcolsep}{6pt}
\renewcommand{\arraystretch}{1.08}
\providecommand{\onsetcell}[2]{\makecell{#1\\[-1pt]{\scriptsize #2 reached}}}
\caption{\textbf{Moshi component ablation on 40 matched CANDOR replays.}
Definitions and reporting follow \cref{tab:main_results}.}
\label{tab:moshi_ablation_results}
\begin{tabular}{@{}lccc@{}}
\toprule
Method & $T_{\mathrm{rep}}$ & $T_{\mathrm{psil}}$ & $T_{\mathrm{gib}}$ \\
\midrule
Baseline & \onsetcell{$19.6\pm7.5$}{$26/40$} & \onsetcell{$9.0\pm0.8$}{$40/40$} & \onsetcell{$13.3\pm0.9$}{$40/40$} \\
IAP & \onsetcell{$24.8\pm21.3$}{$4/40$} & \onsetcell{$33.0\pm14.3$}{$13/40$} & \onsetcell{\NR}{$0/40$} \\
DRP & \onsetcell{\NR}{$0/40$} & \onsetcell{$9.6\pm0.6$}{$37/40$} & \onsetcell{$13.7\pm2.0$}{$40/40$} \\
IAP+DRP & \onsetcell{\NR}{$0/40$} & \onsetcell{$30.3\pm16.0$}{$3/40$} & \onsetcell{\NR}{$0/40$} \\
\bottomrule
\end{tabular}
\end{table}

\subsection{DRP Ablation}
\label{app:drp_ablation}
DRP has two parts (\cref{app:drp}). The first penalizes recent words
(\cref{eq:drp_repetition_penalty}) and then adds an offset that keeps the
total probability of words unchanged (\cref{eq:drp_lexical_mass}). The
second is a \texttt{PAD} safeguard against long silence
(\cref{eq:drp_pad_control}). To see what each part does, we compare them on
40 matched 20-minute PersonaPlex replays. All runs use IAP ($P=1$).

\begin{table}[!htbp]
\centering
\footnotesize
\setlength{\tabcolsep}{5pt}
\caption{\textbf{DRP ablation on 40 matched PersonaPlex CANDOR replays.}
All runs use IAP and last 20 minutes. Rep. and Sil. count runs that reach
repetition or prolonged silence. Output words/min is the number of
model-generated words divided by the full 20-minute replay duration,
including listening and silent periods; it is not an active-speech rate.}
\label{tab:drp_ablation}
\begin{tabular}{@{}lrrr@{}}
\toprule
Method & Rep. $\downarrow$ & Sil. $\downarrow$ &
\makecell{Output\\words/min} \\
\midrule
IAP only                                & $34/40$ & $4/40$  & $158.7$ \\
IAP + standard RP                       & $5/40$  & $21/40$ & $126.9$ \\
IAP + mass-preserving RP                & $2/40$  & $13/40$ & $69.3$ \\
IAP + DRP                               & $0/40$  & $5/40$  & $76.8$ \\
\bottomrule
\end{tabular}
\end{table}

The first row of Table~\ref{tab:drp_ablation} uses IAP alone, without any
repetition penalty, and reaches repetition in 34 of 40 runs. In the second
row, standard RP reduces repetition to 5 runs, but silence increases from 4
to 21 runs, because penalizing words makes \texttt{PAD} relatively more
likely. In the third row, mass-preserving RP keeps the total probability of
words unchanged, so the probability removed from recent words goes to other
words instead of \texttt{PAD}; repetition and silence fall to 2 and 13 runs,
respectively. In the last row, adding the \texttt{PAD} safeguard (full DRP)
further reduces silence to 5 runs, and repetition falls to zero.    DRP also keeps the amount of speech at a normal level. In the first two minutes of the IAP-only runs, before any repetition or silence occurs, the model produces $82.8 \pm 39.1$ words/min (median 81.5). With DRP, it produces 76.8 words/min, a similar amount. IAP alone produces many more words over the full 20
minutes (158.7 words/min), likely because of its repetition loops. In short, these
results show that DRP is well suited to reducing both repetition and
silence.

\subsection{Alternative Repetition Controls}
\label{app:alternative_controls}
We compare DRP with three other ways to suppress recently used words on the
same 40 PersonaPlex replays. The model, input, seed, and IAP ($P=1$) are
fixed. Standard RP applies the CTRL penalty~\citep{keskar2019ctrl} without
DRP's mass preservation or \texttt{PAD} safeguard. LZ~\citep{ginart2026lz}
uses the published settings ($\alpha=0.15$, 512-token lookback, 32-token
buffer). The Bayesian penalty~\citep{fan2026bayesian} needs a prior that says how
often each text token is normally used. To estimate it, we run baseline
PersonaPlex for four minutes on 20 other CANDOR replays (ranks 41--60 by
duration, right after the 40 evaluation replays) and count the text tokens
it produces, with add-one smoothing. These 20 replays do not overlap with
the 40 evaluation replays.

\begin{table}[!htbp]
\centering
\footnotesize
\setlength{\tabcolsep}{5pt}
\caption{\textbf{Alternative repetition controls on 40 matched PersonaPlex
CANDOR replays.} All runs use IAP and last 20 minutes. Rep. and Sil. count
runs that reach repetition or prolonged silence. Output words/min is defined
as in Table~\ref{tab:drp_ablation}.}
\label{tab:alternative_repetition_controls}
\begin{tabular}{@{}lrrr@{}}
\toprule
Control & Rep. $\downarrow$ & Sil. $\downarrow$ & \makecell{Output\\words/min} \\
\midrule
IAP only     & $34/40$ & $4/40$  & $158.7$ \\
\cdashline{1-4}[0.4pt/1.2pt]
Standard RP  & $5/40$  & $21/40$ & $126.9$ \\
LZ           & $31/40$ & $10/40$ & $141.2$ \\
Bayesian     & $0/40$  & $7/40$  & $220.2$ \\
\cdashline{1-4}[0.4pt/1.2pt]
DRP          & $0/40$  & $5/40$  & $76.8$ \\
\bottomrule
\end{tabular}
\end{table}

Under these settings, LZ still reaches repetition in 31 of 40 runs, and
Standard RP produces silence in 21. Bayesian and DRP both prevent the
thresholded repetition event; silence occurs in 7 Bayesian runs and 5 DRP
runs. DRP produces about as much speech as the model does in the first two
minutes of the IAP-only runs, before any repetition or silence occurs
(76.8 vs.\ 82.8 words/min; Appendix~\ref{app:drp_ablation}), whereas
Bayesian produces far more (220.2 words/min). DRP is also simpler to use: it needs only the last $N$ lexical tokens, whereas LZ must keep a compressor state during generation
and Bayesian must first estimate a prior from separate data. We used the published settings for LZ and Bayesian without tuning them for PersonaPlex.

\subsection{Hyperparameter Sensitivity}
\label{app:hyperparameters}
We vary one DRP setting at a time on the same 40 PersonaPlex replays, keeping
the other settings at their final values. \Cref{tab:personaplex_hyperparameters}
shows that the main event counts are stable around the final configuration.
Penalty strengths of $1.3$ and $1.5$ both eliminate repetition. $\rho=1.5$
leaves slightly fewer silent runs (3 vs.\ 5 of 40), but we use $\rho=1.3$ for PersonaPlex throughout the paper. $\rho=2.0$ gives the fewest silent runs (1 of 40), but
its output rate (244.2 words/min) is about three times that of the early
IAP-only runs (82.8 words/min; \cref{app:drp_ablation}), so fewer failure
events do not always mean better speech. Changing the history length $N$ from
256 to 1024 tokens changes the repetition count by at most one run.
Increasing the maximum \texttt{PAD}-logit penalty from 0 to 3 reduces silence
from 13 to 5 runs. We therefore treat the final setting as a fixed operating
point rather than as a uniquely optimal choice.

\begin{table}[!htbp]
\centering
\footnotesize
\setlength{\tabcolsep}{4.5pt}
\caption{\textbf{PersonaPlex DRP hyperparameter sensitivity on 40 matched
20-minute CANDOR replays.} Each block varies one setting from the final configuration ($\rho = 1.3$, $N = 512$, $b_{\max} = 3$). Rep. and Sil. count runs
that reach repetition or prolonged silence. Output words/min is defined as in Table~\ref{tab:drp_ablation}.}
\label{tab:personaplex_hyperparameters}
\begin{tabular}{@{}llrrr@{}}
\toprule
Varied setting & Value & Rep. $\downarrow$ & Sil. $\downarrow$ & \makecell{Output\\words/min} \\
\midrule
\multirow{4}{*}{Penalty $\rho$}
 & $1.2$ & $4/40$ & $4/40$ & $87.2$ \\
 & $1.3$ (final) & $0/40$ & $5/40$ & $76.8$ \\
 & $1.5$ & $0/40$ & $3/40$ & $84.1$ \\
 & $2.0$ & $0/40$ & $1/40$ & $244.2$ \\
\midrule
\multirow{3}{*}{History length $N$}
 & $256$ & $0/40$ & $5/40$ & $76.1$ \\
 & $512$ (final) & $0/40$ & $5/40$ & $76.8$ \\
 & $1024$ & $1/40$ & $4/40$ & $66.6$ \\
\midrule
\multirow{3}{*}{$b_{\max}$}
 & $0$ & $2/40$ & $13/40$ & $69.3$ \\
 & $1$ & $2/40$ & $11/40$ & $71.4$ \\
 & $3$ (final) & $0/40$ & $5/40$ & $76.8$ \\
\bottomrule
\end{tabular}
\end{table}

\subsection{Human Listening Comparison}
\label{app:human_tradeoff}
Because DRP penalizes every recently used word, including words that people
naturally repeat, such as function words, backchannels, and topic words, it
could make speech less natural. We ran a blinded A/B listening study to check
whether DRP makes ordinary speech worse as a conversation goes on. We compare
IAP+DRP with IAP alone on Moshi, isolating the effect of DRP. We keep IAP in both conditions because
without it, both would produce gibberish after about 13 minutes, leaving
nothing meaningful to compare in later phases. The study uses 37 matched
pairs sampled early, middle, and late in the conversation (near 2, 10, and 18
minutes). In each trial, listeners heard 8 seconds of shared conversation
context followed by two 12-second continuations, and chose A, B, or a tie on
five dimensions: naturalness, intelligibility, prosody, conversational style,
and appropriateness to the context. Conditions, transcripts, and generation
times were hidden. Fifteen participants recruited through Prolific, who gave
informed consent and were compensated, completed the study. We report a
preference score for IAP+DRP: choosing it counts as 1, a tie as 0.5, and
choosing IAP alone as 0, so 0.5 means no preference.

Overall, scores are close to 0.5 on every dimension (0.47--0.53;
Table~\ref{tab:human_tradeoff_form1}), but the point estimates shift over time.
Early in the conversation, they favor IAP alone (0.31--0.43), clearly so
for intelligibility and context. In the middle phase, they favor IAP+DRP
(0.56--0.60), clearly so for naturalness. In the late phase, they still
lean toward IAP+DRP (0.53--0.60), but every interval includes 0.5. This
suggests that DRP suppresses some natural word reuse early in a
conversation but may help once repetition starts to build up.

\begin{table}[!htbp]
\centering
\footnotesize
\setlength{\tabcolsep}{3.5pt}
\renewcommand{\arraystretch}{1.12}
\caption{\textbf{Human listening preferences for IAP+DRP over IAP alone on
Moshi.} Each cell shows how often listeners chose IAP+DRP, with ties counted
as half. Higher values mean IAP+DRP was chosen more often, and 0.5 means no
preference. Brackets show 95\% bootstrap intervals; an interval that includes
0.5 indicates no clear preference. Numbers in parentheses are the numbers of
matched pairs; the study contains 285 judgments per dimension from 15
listeners.}
\label{tab:human_tradeoff_form1}
\resizebox{\linewidth}{!}{%
\begin{tabular}{@{}lccccc@{}}
\toprule
Phase & Naturalness & Intelligibility & Prosody & Conv.\ style & Context \\
\midrule
Early (14)  & 0.43 [0.30, 0.55] & 0.38 [0.27, 0.48] & 0.41 [0.28, 0.54] & 0.43 [0.30, 0.56] & 0.31 [0.22, 0.41] \\
Middle (11) & 0.60 [0.51, 0.69] & 0.59 [0.50, 0.68] & 0.60 [0.50, 0.71] & 0.60 [0.50, 0.71] & 0.56 [0.46, 0.68] \\
Late (12)   & 0.53 [0.43, 0.65] & 0.55 [0.41, 0.69] & 0.60 [0.41, 0.78] & 0.58 [0.43, 0.73] & 0.58 [0.44, 0.73] \\
\midrule
All (37)    & 0.52 [0.44, 0.59] & 0.50 [0.43, 0.57] & 0.53 [0.45, 0.61] & 0.53 [0.45, 0.61] & 0.47 [0.40, 0.55] \\
\bottomrule
\end{tabular}%
}
\end{table}

\subsection{Computational Cost}
\label{app:efficiency}
\label{sec:supp-computational-efficiency}

\paragraph{Setup.}
We measure the inference cost of each method on PersonaPlex using a single
NVIDIA RTX A6000 GPU with bfloat16 weights. We compare the standard FIFO
cache, a StreamingLLM-style cache that keeps the first four entries and
re-indexes the whole cache at every step, DRP, IAP, and IAP+DRP. All methods
use the same 440-second replay, persona, sampling settings, seed, and
3,000-entry cache. Timing starts only after the cache is full, since the methods differ only once old entries begin to be evicted. We then measure 1,000 consecutive frames (80 seconds of audio) and report the average time per frame. Each method is run five times,
each in a new process. Model loading, file I/O, and resampling are excluded.

\begin{center}
\begin{minipage}{0.8\linewidth}
\centering
\small
\captionsetup{hypcap=false}
\setlength{\tabcolsep}{4pt}
\captionof{table}{\textbf{Per-frame inference cost on PersonaPlex.}
Mean${}\pm{}$SD over five runs with a full 3,000-entry cache. Overhead is the
increase in total time relative to FIFO. An RTF below 1 means faster than
real time.}
\label{tab:computational-efficiency}
\resizebox{\linewidth}{!}{%
\begin{tabular}{@{}lrrrr@{}}
\toprule
Method & \makecell{LM GPU\\(ms/frame)} & \makecell{Total\\(ms/frame)} & RTF & Overhead \\
\midrule
FIFO & $39.939{\pm}0.111$ & $47.653{\pm}0.163$ & $0.5957{\pm}0.0020$ & -- \\
StreamingLLM-style & $74.818{\pm}0.153$ & $82.583{\pm}0.229$ & $1.0323{\pm}0.0029$ & $+73.3\%$ \\
DRP & $40.485{\pm}0.100$ & $48.944{\pm}0.138$ & $0.6118{\pm}0.0017$ & $+2.7\%$ \\
IAP & $51.153{\pm}0.136$ & $59.083{\pm}0.598$ & $0.7385{\pm}0.0075$ & $+24.0\%$ \\
IAP+DRP & $51.694{\pm}0.136$ & $60.198{\pm}0.185$ & $0.7525{\pm}0.0023$ & $+26.3\%$ \\
\bottomrule
\end{tabular}%
}
\end{minipage}
\end{center}

\paragraph{Metrics.}
\Cref{tab:computational-efficiency} reports the time per frame.
\emph{LM GPU} is the time the language model takes on the GPU to produce one frame.
\emph{Total} is the time of the full per-frame pipeline, including input-audio
encoding, output-audio decoding, and DRP's logit processing. \emph{RTF} is the
time, in seconds, needed to process one second of audio; an RTF below 1 means
the model runs in real time.

\paragraph{Results.}
DRP adds almost no cost (+2.7\%), since it only edits the text logits. IAP
adds 24.0\%, but IAP+DRP still runs well within real time (RTF 0.75).
StreamingLLM-style is the slowest (+73.3\%): because it re-indexes every
cached entry at each step, its RTF exceeds 1, so it cannot keep up with real
time.
\FloatBarrier

\section{Exploratory Observations and Limitations}
\label{app:exploratory}
This section reports exploratory observations outside the main benchmark and
discusses the limitations of this work. \Cref{app:half_duplex} checks whether
repetition also appears in additional speech models; because it uses
different protocols, its results are not pooled with FD-LHB.
\Cref{app:limitations} then discusses the limitations.

\subsection{Exploratory Observations on Additional Speech Models}
\label{app:half_duplex}
We also ran several additional speech models on long inputs to see whether
repetition appears beyond the models in our main benchmark
(\cref{tab:half_duplex_behavior}).
Each model uses its own prompt, input, context unit, and stopping rule, and we
observe one sequence per model. These results are therefore qualitative: they
cannot be compared with FD-LHB or used to rank the models. Seven of the nine
models eventually repeat, while GLM-4-Voice and LLaMA-Omni2 stay stable in the
tested sequence.

\begin{table}[!htbp]
\centering
\footnotesize
\setlength{\tabcolsep}{4pt}
\caption{\textbf{Exploratory observations on additional speech models.}
One sequence per model; lengths and protocols differ across models.}
\label{tab:half_duplex_behavior}
\begin{tabularx}{\linewidth}{@{}l p{0.22\linewidth} X@{}}
\toprule
Model & Tested length & Observed behavior \\
\midrule
mini-omni~\citep{xie2024mini} & 2,048 tokens & Repeats replies before reaching the context limit. \\
mini-omni2~\citep{xie2024mini2} & 2,048 tokens & Coherent until near the limit, then repeats. \\
Covo-Audio~\citep{wang2026covo} & a few thousand tokens & Replies converge to a small repeated set. \\
Kimi-Audio~\citep{ding2025kimi} & about 1.3K of 8K tokens & Switches to describing the audio and repeats that description. \\
Qwen2.5-Omni~\citep{xu2025qwen25omnitechnicalreport} & about 11 turns / 2.5K tokens & Outputs template text, then repeats one response. \\
Step-Audio 2-mini~\citep{wu2025step} & tested dialogue & Coherent early, then repeats replies. \\
GLM-4-Voice~\citep{zeng2024glm} & 8,192 tokens & Stable and on-topic. \\
LLaMA-Omni2~\citep{fang2025llama} & 60 turns / 3,232 tokens & Stable. \\
Baichuan-Audio~\citep{li2025baichuan} & separate long-input test & Repeats one answer; the run later ends when GPU memory runs out. \\
\bottomrule
\end{tabularx}
\end{table}

\subsection{Limitations}
\label{app:limitations}
\label{sec:limitations}

\paragraph{Scope of the diagnosis.}
Our diagnosis covers only the two sliding-window models, Moshi and
PersonaPlex. Growing-context models such as BayLing-Duplex also collapse
before reaching their context limit, but we do not identify the cause, and IAP
does not apply to them because they never evict cache entries. Even for the sliding-window models, our evidence for Initialization-Anchor Eviction comes mainly from controlled experiments. The fixed-slot test (\cref{sec:diagnosis_firstentry}) shows that the first KV entry is important, while the matched-cache test (\cref{sec:state_drift}) shows that the previous cache state affects the failure.
The fixed-slot test measures predictive entropy, so it does not directly measure dialogue quality. It also does not show how often the same failure occurs across different conversations or inputs.
Broader tests across
models, checkpoints, inputs, random seeds, architectures, and languages are
needed.

\paragraph{Evaluation horizon and cross-model comparison.}
FD-LHB observes each model for a finite time, so \textsc{NR} means only that a failure did not occur before the run ended, not that the model would remain stable indefinitely. The evaluated models also differ in architecture, context management, and generation settings, including model-specific text temperatures in the adaptive track. Therefore, comparisons between methods within the same model are more directly interpretable than comparisons across different models.

\paragraph{Metrics and trade-offs.}
Our evaluation metrics are not perfect measures of overall conversation quality. First, our criteria (\cref{sec:failure_detection}) detect repetition, silence, and gibberish separately. A conversation that avoids these failures may still be incoherent or irrelevant. Second, reducing one type of failure can introduce a trade-off elsewhere. For example, DRP may penalize natural word reuse, and listeners preferred IAP alone over IAP+DRP early in the conversation (\cref{app:human_tradeoff}). Moreover, the proposed methods do not eliminate all failures: in PersonaPlex, IAP+DRP still produces a two-minute silence in 19 of 40 replays. Finally, the Gemini-based gibberish detector is also imperfect. Our human validation (\cref{app:human_validation}) found no false positives but did find some missed cases, suggesting that the reported gibberish rates may underestimate the true frequency. These limitations motivate broader and longer-term human evaluation of both conversation quality and possible trade-offs.

\end{document}